\documentclass[a4paper,fleqn]{cas-dc}

\usepackage[numbers]{natbib}
\def\tsc#1{\csdef{#1}{\textsc{\lowercase{#1}}\xspace}}
\tsc{WGM}
\tsc{QE}
\begin{document}
\let\WriteBookmarks\relax
\def\floatpagepagefraction{1}
\def\textpagefraction{.001}

\shorttitle{Improving LLM-Generated Process Model Quality Through Reinforcement Learning}

\shortauthors{}  

\title [mode = title]{Improving LLM-Generated Process Model Quality Through Reinforcement Learning: The Role of Reward Function Design}  




\author[1,2]{Alexander Rombach}[orcid=0000-0002-9173-4215]
\cormark[1]
\ead{alexander_michael.rombach@uni-saarland.de}
\credit{Conceptualization, Methodology, Software, Writing – original draft, Writing – review and editing}

\author[1,2]{Chantale Lauer}%[<options>]
\ead{chantale.lauer@dfki.de}
\credit{Conceptualization, Data curation, Software, Writing – original draft, Writing – review and editing}

\author[1,2]{Nijat Mehdiyev}[orcid=0000-0001-7899-1017]
\ead{nijat.mehdiyev@dfki.de}
\credit{Supervision, Methodology, Formal analysis, Validation, Writing – original draft, Writing – review and editing}

\affiliation[1]{organization={German Research Center for Artificial Intelligence (DFKI)},
            addressline={Campus D3 2}, 
            city={Saarbrücken},
            postcode={66123}, 
            state={Saarland},
            country={Germany}}

\affiliation[2]{organization={Saarland University},
            addressline={Campus D3 2}, 
            city={Saarbrücken},
            postcode={66123}, 
            state={Saarland},
            country={Germany}}

\cortext[1]{Corresponding author}



\begin{abstract}
Large language models (LLMs) can generate BPMN process models from natural-language descriptions, yet supervised fine-tuning (SFT) limits their output quality to the patterns present in the training data. Reinforcement learning (RL) can optimize beyond this ceiling using external quality measures, but how the reward function should be designed when quality is multi-dimensional remains unexplored. We present a systematic investigation of reward function design for RL-based process model generation, training two LLM families (Llama~3.1 8B, Qwen~2.5 14B) under 48 configurations using Group Sequence Policy Optimization with rewards derived from an automated evaluation framework comprising 38 metrics across syntactic, pragmatic, and semantic quality. Three findings emerge. First, RL significantly improves pragmatic and syntactic quality while preserving semantic fidelity, reducing output variability by more than sixfold. Second, equal reward weighting consistently outperforms targeted weighting: emphasizing a specific dimension fails to improve it and can collapse the model into a low-quality mode. Third, design choices interact with model architecture in non-trivial ways: the invalidity penalty is essential for one model but irrelevant for the other, and SFT initialization is indispensable for one architecture but counterproductive for another. These results demonstrate that reward composition is a primary determinant of optimization outcomes, with effects as large as the decision to apply RL itself. The findings generalize to any structured generation task where quality is assessed along multiple automated dimensions. We release our implementation and experimental code at \url{https://github.com/chlauer99/RL_for_process_modeling}.

\end{abstract}


\iffalse
\begin{highlights}
\item Presents GSPO-based RL framework for LLM-driven BPMN generation using quality rewards.
\item Offers transferable insights for structured generation and reward function design.
\item Reveals reward-architecture-initialization interactions and offers design guidelines.
%\item Verifiable domain rewards replace learned proxy models, improving RL transparency.
\item Reproducible benchmark: 48 configs across two LLM families for RL process generation.
\end{highlights}
\fi

%\nocite{*}

% Keywords

\maketitle

% Main text
%\section{}\label{}

% Use the style of numbering in square brackets.
% If nothing is used, default style will be taken.
%\begin{enumerate}[a)]
%\item 
%\item 
%\item 
%\end{enumerate}  

%\begin{itemize}
%\item 
%\item 
%\item 
%\end{itemize}  

%\begin{description}
%\item[]
%\item[] 
%\item[] 
%\end{description}  

\section{Introduction}
\label{sec:introduction}

The creation of Business Process Model and Notation (BPMN) models is a complex task requiring both domain knowledge and proficiency in modeling conventions. As organizations increasingly seek to document, analyze, and automate their workflows, the demand for process models has outpaced the availability of trained modelers, creating a bottleneck that limits the scalability of business process management initiatives~\cite{kampik2025large}. Large language models (LLMs) offer a promising path to addressing this bottleneck by enabling the generation of process models directly from natural-language descriptions, thereby lowering the expertise barrier and accelerating the modeling lifecycle~\cite{klievtsova2023conversational}.

Recent work has demonstrated that LLMs can indeed produce BPMN-like structures from text, with interactive systems supporting iterative refinement~\cite{kourani2024promoai,lauer_conversational_modeling} and compact intermediate representations improving generation robustness~\cite{brissard2025representation,kopke2025efficient}. Systematic benchmarking has further shown that open-source LLMs achieve competitive syntactic and pragmatic quality on BPMN generation tasks, though semantic fidelity and output validity remain significant challenges~\cite{lauer2026bef4llm}. Supervised fine-tuning (SFT) with parameter-efficient methods such as LoRA~\cite{hu2022lora} has been shown to improve structural correctness and validity, enabling smaller models to outperform larger untuned baselines. However, SFT is inherently limited by the quality and diversity of the training data it imitates: it can teach a model to reproduce the patterns present in the training corpus but cannot drive the model beyond this ceiling toward outputs that are \emph{better} than the demonstrations.

Reinforcement learning (RL) offers a fundamentally different training signal. Rather than imitating reference outputs, the model is optimized against an external quality measure that evaluates its generated outputs directly. The recent emergence of reinforcement learning with verifiable rewards (RLVR)~\cite{lambert2024tulu3,guo2025deepseekr1} has demonstrated that when automated, deterministic evaluation is available (as in mathematical reasoning or code generation), RL can produce substantial improvements over SFT baselines without requiring human preference data. Process model generation is uniquely suited to this paradigm: the BEF4LLM evaluation framework~\cite{lauer2026bef4llm} provides 38 automated metrics across three established quality dimensions (syntactic, pragmatic, and semantic quality) that can serve directly as reward signals, enabling RLVR for process modeling.

However, the design of the reward function itself, specifically how these multi-dimensional quality metrics are composed into a training signal, has received almost no systematic attention. Existing work on RL for process modeling~\cite{berti2025specializing} demonstrates that RL improves generation quality but uses a single reward formulation without investigating how the composition of the reward affects optimization outcomes. More broadly, most RLVR applications operate with single-dimensional rewards (correct/incorrect for mathematics, tests passed/failed for code), whereas process model quality is inherently multi-dimensional: syntactic conformance, pragmatic comprehensibility, and semantic fidelity represent distinct and potentially competing optimization objectives. How these objectives should be balanced, whether dimensional emphasis through reward weighting can steer optimization toward desired quality profiles, and how design choices such as invalidity penalties interact with model architecture are questions that remain unanswered.

This paper addresses this gap through a systematic empirical investigation of reward function design for RL-based process model generation. We employ Group Sequence Policy Optimization (GSPO)~\cite{zheng2025gspo}, a sequence-level RL method that aligns naturally with holistic quality evaluation, and train two open-source LLM families (Llama~3.1 8B and Qwen~2.5 14B) under 48 experimental configurations spanning six reward functions, two base model initialization strategies, and two decoding constraints. The reward functions vary along two primary design axes: dimensional
weighting (equal vs.\ targeted emphasis) and invalidity penalty (negative
penalty vs.\ zero). In addition, we include two mathematically equivalent
equal-weight implementations, $R_{\mathrm{avg}}$ and~$R_1$, to check
whether implementation pathway differences produce measurable effects. All configurations are evaluated using the BEF4LLM framework and compared through paired permutation tests~\cite{good2005permutation} with Bonferroni correction.

The study is guided by three research questions. RQ1 asks whether GSPO with domain-grounded rewards improves process model quality beyond SFT. RQ2 investigates how reward function composition, specifically dimensional weighting and the invalidity penalty, affects the balance across quality dimensions. RQ3 examines how the choice of base model initialization (SFT-initialized vs.\ untrained, i.e., BPM-unadapted instruction-tuned base) affects RL optimization outcomes.

The results yield several findings with implications beyond the BPMN domain. First, GSPO significantly improves pragmatic and syntactic quality for both model families while largely preserving semantic quality, with only a small positive effect for Llama and no significant effect for Qwen. The most pronounced effect is a dramatic reduction in output variability: RL-trained models produce substantially more consistent outputs than SFT-only baselines. Second, and most counter-intuitively, equal reward weighting outperforms all targeted weighting schemes. Emphasizing a specific quality dimension in the reward not only fails to improve that dimension but can cause the model to collapse into a narrow, low-quality mode. Third, the invalidity penalty plays a role far beyond discouraging invalid outputs. For Qwen, it acts as an implicit diversity regularizer that reduces the risk of template convergence, while for Llama it has no measurable effect. Fourth, the necessity of SFT initialization is architecture-dependent: indispensable for Llama under our training configuration, largely redundant for Qwen, and associated with lower semantic quality for Qwen after RL. These findings demonstrate that reward function design for multi-dimensional structured generation is a non-trivial engineering challenge where seemingly minor choices produce effects as large as the decision to apply RL in the first place.

The contributions of this paper are as follows:
\begin{enumerate}
    \item A training framework for applying GSPO either after SFT or directly to BPM-unadapted instruction-tuned base models, using domain-grounded, multi-dimensional reward signals derived from the BEF4LLM quality framework for LLM-based BPMN process model generation.

    \item A systematic empirical investigation of reward function design for structured generation, covering dimensional weighting, invalidity penalty, and the exploratory comparison of equivalent reward implementation pathways across two LLM architectures and two initialization strategies, comprising 48 experimental configurations evaluated on 105 process descriptions.

    \item An analysis of the interaction effects between reward design, model architecture, and initialization strategy, revealing that these factors are not independently tunable but interact in non-trivial ways that require empirical evaluation rather than principled defaults.

    \item Practical guidelines for applying RL to multi dimensional structured generation tasks, including the finding that equal weighting is the most robust default in our experiments, the penalty regime should match the base model's validity behavior, and SFT initialization should be informed by the base model's task-relevant prior capabilities.
\end{enumerate}

The remainder of this paper is structured as follows. Section~\ref{sec:rw_rl} reviews related work on LLMs in BPM, RL for LLMs, quality assessment frameworks, and RL for structured generation. Section~\ref{sec:method} describes the methodology, including the training pipeline, quality framework, and reward function design. Section~\ref{sec:experiment} details the experimental setup. Section~\ref{sec:results} presents the results organized by research question. Section~\ref{sec:discussion} discusses the findings, compares with related work, and addresses limitations and future work. Section~\ref{sec:conclusion} concludes the paper.
\section{Related Work}
\label{sec:related}

\iffalse
This section reviews four streams of research that converge in the present study: the emerging role of LLMs in process modeling (Section~\ref{sec:rw_bpm}), reinforcement learning methods for LLMs with an emphasis on verifiable rewards (Section~\ref{sec:rw_rl}), quality assessment frameworks that enable automated evaluation (Section~\ref{sec:rw_quality}), and the nascent field of RL for structured generation tasks (Section~\ref{sec:rw_structured}).
\fi
%--------------------------------------------------------------
\subsection{LLMs in Business Process Modeling}
\label{sec:rw_bpm}

The application of LLMs to business process management (BPM) has advanced rapidly from proof-of-concept demonstrations to systematic evaluations and specialized training pipelines. Early work established that LLMs can translate natural-language descriptions into formal process representations~\cite{klievtsova2023conversational}, and interactive systems such as ProMoAI~\cite{kourani2024promoai} demonstrated iterative refinement capabilities. A recent literature review highlights the methodological shift from non-generative extraction pipelines to end-to-end generative approaches, identifying domain adaptation through fine-tuning and intermediate representation design as the two most active frontiers ~\cite{kampik2025large}.

Systematic benchmarking has revealed both the promise and the limitations of current LLMs. \citet{lauer2026bef4llm} introduce BEF4LLM, a framework comprising 38 metrics across syntactic, pragmatic, and semantic quality dimensions, and benchmark 17 open-source LLMs on BPMN generation---finding that LLMs achieve competitive structural quality but lag behind human experts on semantic fidelity. \citet{horner2026bpmn} independently confirm these patterns while proposing an automated BPMN~2.0 generation pipeline with quality assessment. \cite{lauer2026humancentered} conducted a human-centered study and similarly identified semantic quality of process models as a key factor influencing perceived usability and trust, which were negatively affected by low semantic quality of the generated process models.
\citet{celikmasat2025bpmn} demonstrate that instruction-tuned open-source LLMs can be further improved through supervised fine-tuning (SFT) on curated process model datasets.

The choice of output representation has emerged as a critical factor. Compact intermediate languages based on JSON reduce token counts substantially compared to XML while preserving BPMN semantics, enabling efficient fine-tuning and improving generation robustness~\cite{brissard2025representation,kopke2025efficient}. The present study builds on this foundation---using a compact JSON-based process representation as the output format and BEF4LLM as the evaluation framework---but shifts the focus from representation and benchmarking to the \emph{training methodology}, specifically the design of RL reward functions for multi-dimensional quality optimization.

%--------------------------------------------------------------
\subsection{Reinforcement Learning for Large Language Models}
\label{sec:rw_rl}

Reinforcement learning (RL) has become central to post-training alignment of LLMs. The standard RLHF pipeline \cite{ouyang2022training} fine-tunes a pre-trained model using a learned reward model and Proximal Policy Optimization (PPO)~\cite{schulman2017ppo}, constrained by a KL-divergence penalty. While effective for preference alignment, PPO-based methods suffer from instability, high computational overhead, and sensitivity to reward scaling---issues that become more pronounced in long-horizon structured generation tasks.

Direct Preference Optimization (DPO)~\cite{rafailov2023dpo} sidesteps the explicit reward model by reparameterizing the RLHF objective to operate directly on preference pairs. While DPO simplifies training, it requires paired preference data and does not naturally accommodate the kind of verifiable, metric-based reward signals available in our setting. Kahneman-Tversky Optimization (KTO)~\cite{ethayarajh2024kto} further relaxes data requirements to unpaired signals but shares the same fundamental limitation.

The emergence of \emph{reinforcement learning with verifiable rewards} (RLVR)~\cite{lambert2024tulu3} represents a paradigm shift particularly relevant to our work. Rather than learning reward models from human preferences, RLVR leverages automated, deterministic evaluation, such as mathematical correctness verification or code test-case execution, as reward signals. DeepSeek-R1~\cite{guo2025deepseekr1} demonstrated that RLVR with Group Relative Policy Optimization (GRPO)~\cite{shao2024deepseekmath} can unlock sophisticated reasoning capabilities without any supervised demonstration data, sparking widespread interest in the approach. GRPO samples multiple candidate outputs per prompt, normalizes rewards within each group, and updates the policy to favor relatively better responses---reducing sensitivity to reward miscalibration and eliminating the need for a separate critic network.

Group Sequence Policy Optimization (GSPO)~\cite{zheng2025gspo} adapts this paradigm by operating at the level of complete sequences rather than individual tokens. For each prompt, multiple candidates are sampled, scored holistically, and ranked; the model is updated to increase the likelihood of higher-ranked sequences. This sequence-level formulation aligns naturally with evaluation frameworks that assess complete outputs such as process model quality metrics that can only be computed for entire models. The present study adopts GSPO as the RL algorithm, motivated by this alignment between sequence-level optimization and holistic quality assessment.

%--------------------------------------------------------------
\subsection{Quality Assessment as Reward Signal}
\label{sec:rw_quality}

%The use of automated quality metrics as RL reward signals connects two traditionally separate research streams: process model quality assessment from BPM and reward function design from RL.

Process model quality has been studied through the lens of the semiotic quality framework~\cite{lindland1994quality,krogstie2006quality}, operationalized by~\citet{reijers2010siq} as the SIQ framework distinguishing syntactic, semantic, and pragmatic quality. Concrete metrics include structural complexity measures~\cite{mendling2008metrics,vanderfeesten2008quest}, threshold-based pragmatic scoring~\cite{sanchez2011thresholds,boomsma2009thresholds}, and graph-structural and behavioral similarity measures for semantic quality~\cite{dijkman2011similarity,vandongen2013similarity}. The BEF4LLM framework~\cite{lauer2026bef4llm} synthesizes these into a unified pipeline producing normalized per-dimension scores---a property that makes them directly usable as reward components.

From the RL perspective, reward function design is a long-standing challenge. Reward shaping~\cite{ng1999reward} and multi-objective reward decomposition~\cite{hayes2022multi} address the problem of guiding agents toward desired behaviors when the objective is complex or multi-faceted. In the LLM context, most reward functions are either binary (correct/incorrect, as in RLVR for mathematics) or scalar (a single preference score from a reward model). Multi-dimensional reward functions---where separate components measure distinct quality aspects that must be balanced---remain underexplored. Our work contributes to this gap by systematically investigating how the composition of a multi-dimensional quality reward (weighting, penalty, aggregation) affects optimization outcomes in structured generation.

%--------------------------------------------------------------
\subsection{RL for Structured Generation}
\label{sec:rw_structured}

%The application of RL specifically to structured text generation---where outputs must satisfy formal syntactic constraints, adhere to schema specifications, and preserve semantic content---represents a nascent but growing research direction that bridges the streams reviewed above.

In the BPM domain, \citet{berti2025specializing} apply RLVR to specialize LLMs for process modeling using Partially Ordered Workflow Language (POWL) as the target representation. Their approach combines structural and behavioral reward signals with LLM-as-a-judge feedback, demonstrating improvements over both pre-trained and SFT baselines. While their work establishes the viability of RL for process modeling, it uses a single reward formulation without investigating how reward composition affects optimization dynamics. Specifically, they do not examine dimensional weighting, penalty structure, or the interaction between reward design and model architecture---the central questions of the present study.

Beyond BPM, \citet{hu2024rlstruct} propose RL-Struct, a lightweight RL framework for reliable structured LLM output that decomposes the task into a hierarchy of constraints---structural integrity, format correctness, content accuracy, and validity---optimized through GRPO. Their finding that models exhibit a ``self-paced curriculum,'' sequentially acquiring syntactic proficiency before semantic accuracy, resonates with our observation of differential optimization accessibility across quality dimensions. In code generation, CodeRL~\cite{le2022coderl} uses unit-test pass rates as automated rewards, and subsequent work extends this with curriculum strategies and multi-candidate selection. In mathematical reasoning, DeepSeek-R1~\cite{guo2025deepseekr1} demonstrates that correctness-based RLVR can produce sophisticated problem-solving behavior. These domains share a key property with our setting: the availability of objective, automated evaluation that eliminates the need for human preference data.

However, a critical difference distinguishes our work: process model quality is inherently \emph{multi-dimensional}, requiring reward functions that balance competing objectives. Mathematical reasoning and code generation typically operate with single-dimensional rewards, whereas our three quality dimensions can be in tension with each other. The present study addresses this gap by conducting the first systematic investigation of multi-dimensional reward function design for structured generation, varying the weighting, penalty structure, and aggregation method while analyzing interactions with model architecture and initialization strategy.
\section{Methodology}
\label{sec:method}

This section describes the proposed approach for improving LLM-based process model generation through reinforcement learning. We first define the generation task and its constraints (Section~\ref{method:task}), then introduce the training pipeline (Section~\ref{method:pipeline}). Section~\ref{method:quality} defines the multi-dimensional quality assessment that underpins the reward signals. Section~\ref{method:reward} formally specifies the family of reward functions, Section~\ref{method:rationale} discusses the design rationale, and Section~\ref{method:grammar} introduces grammar-constrained decoding as an additional experimental factor. Table~\ref{tab:notation} summarizes the notation used throughout.

\begin{table}[hbpt]
\centering
\caption{Summary of notation.}
\label{tab:notation}
\smallskip
\begin{tabular}{l p{0.55\columnwidth}}
\toprule
Symbol & Meaning \\
\midrule
$c = (s, x)$ & Prompt: system instruction $s$ and process description $x$ \\
$\tilde{y}$ & Raw output sequence generated by the LLM \\
$y$ & Parsed intermediate process model (from $\tilde{y}$ when parsing succeeds) \\
$\bot$ & Parse or translation failure \\
$\pi_\theta$ & LLM policy with parameters $\theta$ \\
$\phi\colon \mathcal{Y} \rightarrow \mathcal{Y}_{\mathrm{XML}} \cup \{\bot\}$ & Deterministic translator; returns BPMN~XML or $\bot$ on failure \\
$v(\tilde{y})$ & Validity indicator (binary) \\
$r_{\mathrm{syn}},\, r_{\mathrm{pra}},\, r_{\mathrm{sem}}$ & Quality scores (syntactic, pragmatic, semantic) \\
$R(\mathbf{w}, p;\, \tilde{y})$ & Composite reward \\
\bottomrule
\end{tabular}
\end{table}

%--------------------------------------------------------------
\subsection{Task Definition}
\label{method:task}

We address the \emph{text-to-model generation task}, in which a LLM translates a natural-language process description into a structured process model. The input is a prompt $c = (s, x)$, where $s$ is a system instruction defining the modeling task and the expected output format, and $x \in \mathcal{X}$ is a textual process description. Given~$c$, the LLM policy $\pi_\theta$ generates a raw output sequence:
\begin{equation}
\label{eq:generation}
\tilde{y} \sim \pi_\theta(\cdot \mid c)
\end{equation}
where $\tilde{y} \in \tilde{\mathcal{Y}}$ is a token sequence that may or may not constitute a well-formed process model. A parser attempts to map $\tilde{y}$ to a parsed intermediate process model $y \in \mathcal{Y}$, expressed in a compact, JSON-based representation that preserves the organizational and behavioral semantics of BPMN collaborations, including pools, lanes, typed flow nodes, sequence flows, and message flows. If parsing succeeds, a deterministic translator $\phi\colon \mathcal{Y} \rightarrow \mathcal{Y}_{\mathrm{XML}} \cup \{\bot\}$ attempts to map $y$ to a corresponding BPMN~XML file, enabling evaluation and rendering by standard BPMN toolchains; the translator returns~$\bot$ if the mapping fails. If parsing itself fails (due to malformed JSON, structural violations, or incomplete output), the parser likewise returns~$\bot$, indicating an invalid generation attempt.

This formalization reflects a key property of the setting: unlike conventional text generation where any token sequence is a valid output, process model generation introduces a chain of potential failure points (malformed text $\rightarrow$ parse failure $\rightarrow$ translation failure $\rightarrow$ schema violation), all of which must be handled by the evaluation and reward machinery.

The task imposes constraints at multiple levels of abstraction. At the \emph{syntactic level}, the output must conform to the structural rules of the BPMN modeling language: For example, every split gateway must have a matching join gateway, and sequence flows may only connect elements within the same pool. At the \emph{pragmatic level}, the resulting model should be comprehensible to human readers, avoiding unnecessary complexity in size, density, or control-flow structure. At the \emph{semantic level}, the generated model must faithfully capture the activities, control-flow logic, participant roles, and inter-process communication described in~$x$. These constraints distinguish the task fundamentally from open-ended text generation: outputs are structured, multi-dimensional artifacts that must simultaneously satisfy verifiable correctness criteria across all three levels.

%--------------------------------------------------------------
\subsection{Training Pipeline}
\label{method:pipeline}

We investigate a training pipeline that combines SFT with RL, following the general paradigm established in related literature~\cite{ouyang2022training}. The pipeline consists of two stages: SFT provides a behavioral initialization, and GSPO subsequently optimizes for quality objectives that are difficult to express as token-level supervision. Critically, the SFT stage is treated as an experimental variable rather than an invariant: GSPO is applied either to the SFT-initialized LLM or directly to the instruction-tuned base LLM, depending on the experimental condition. This design allows us to isolate the effect of supervised initialization on RL optimization outcomes (RQ3).

\paragraph{Stage~1: Supervised Fine-Tuning (SFT).}
In the first stage, the base LLM is fine-tuned on a corpus of input-output pairs $\mathcal{D}_{\mathrm{SFT}} = \{(c_i, y_i)\}_{i=1}^{N}$, where $c_i = (s_i, x_i)$ is the prompt and $y_i$ is the corresponding target process model. For training, each target process model~$y_i$ is serialized into the JSON-based output format, and the loss is computed over the tokens of this serialization. The model is trained to minimize the per-example average token loss, computed solely on the output tokens:
\begin{equation}
\label{eq:sft}
\mathcal{L}_{\mathrm{SFT}}(\theta)
  = -\frac{1}{N}\sum_{i=1}^{N} \frac{1}{|y_i|}\sum_{t=1}^{|y_i|}
    \log\, p_\theta\!\bigl(y_i^{(t)} \mid c_i,\, y_i^{(<t)}\bigr)
\end{equation}
where $\theta$ denotes the model parameters. This stage teaches the model the mapping from natural-language descriptions to structured process models, establishing familiarity with the output format and basic structural competence. However, SFT optimizes through imitation at the token level: the training signal rewards the model for reproducing reference outputs token by token but does not directly optimize for downstream quality properties of the resulting process model as a whole. A model trained only with SFT may learn to produce outputs that closely resemble training targets at the surface level without reliably satisfying the syntactic rules, pragmatic guidelines, or semantic constraints that collectively determine whether a process model is correct and useful. This fundamental limitation motivates the second training stage.

\paragraph{Stage~2: Reinforcement Learning with GSPO.}
To move beyond imitation toward quality-aware optimization, we apply GSPO~\cite{zheng2025gspo}. 
%GSPO belongs to a family of group-relative RL methods proposed as alternatives to Proximal Policy Optimization (PPO) for language model training. Whereas PPO requires a separately trained value function and applies token-level policy updates constrained by a KL-divergence penalty, GSPO operates at the level of complete generated sequences, treating each full response as a single action. 
As mentioned in \ref{sec:rw_rl}, this formulation aligns naturally with our setting, where reward signals that are derived from process model quality metrics evaluate complete BPMN models rather than intermediate token predictions.

The GSPO training procedure operates as follows. For each training prompt~$c_i$, the current policy~$\pi_\theta$ generates a group of~$K$ candidate output sequences through sampling:
\begin{equation}
\label{eq:gspo_sample}
\tilde{y}_i^{(k)} \sim \pi_\theta(\cdot \mid c_i), \quad k = 1, \ldots, K
\end{equation}
Each candidate is evaluated by the reward function~$R(\cdot)$ (defined in Section~\ref{method:reward}), yielding scalar rewards $\{r_i^{(1)}, \ldots, r_i^{(K)}\}$. Rather than using these as absolute optimization targets, GSPO normalizes them within the group to compute relative advantages:
\begin{equation}
\label{eq:gspo_advantage}
\hat{A}_i^{(k)}
  = \frac{r_i^{(k)} - \mu_i}{\sigma_i + \epsilon}
\end{equation}
where $\mu_i$ and $\sigma_i$ are the mean and standard deviation of the group rewards for prompt~$c_i$, and $\epsilon$ is a small constant for numerical stability. The policy is updated via a clipped surrogate objective. Let $\theta_{\mathrm{old}}$ denote the policy parameters at the time of candidate generation. The length-normalized importance-sampling ratio for the $k$-th candidate of prompt~$i$ is:
\begin{equation}
\label{eq:gspo_ratio}
\rho_i^{(k)}(\theta)
  = \exp\!\left[
    \frac{1}{|\tilde{y}_i^{(k)}|}
    \left(
    \log \pi_\theta\!\bigl(\tilde{y}_i^{(k)} \mid c_i\bigr)
    - \log \pi_{\theta_{\mathrm{old}}}\!\bigl(\tilde{y}_i^{(k)} \mid c_i\bigr)
    \right)
  \right]
\end{equation}
Length normalization ensures that the ratio remains within a comparable numerical range regardless of output length, preventing longer sequences from producing more extreme likelihood ratios~\cite{zheng2025gspo}. The sequence log-probability is computed autoregressively as:
\begin{equation}
\label{eq:gspo_logprob}
\log \pi_\theta\!\bigl(\tilde{y}_i^{(k)} \mid c_i\bigr)
  = \sum_{t=1}^{|\tilde{y}_i^{(k)}|}
    \log \pi_\theta\!\bigl(\tilde{y}_{i,t}^{(k)} \mid c_i,\, \tilde{y}_{i,<t}^{(k)}\bigr)
\end{equation}
The GSPO objective maximizes the clipped surrogate:
\begin{equation}
\label{eq:gspo_objective}
\begin{aligned}
\mathcal{J}_{\mathrm{GSPO}}(\theta)
&= \frac{1}{NK} \sum_{i=1}^{N} \sum_{k=1}^{K}
   \min\!\Bigl(
      \rho_i^{(k)}(\theta)\, \hat{A}_i^{(k)}, \\
&\qquad
      \mathrm{clip}\!\left(
        \rho_i^{(k)}(\theta),
        1{-}\epsilon_c,
        1{+}\epsilon_c
      \right)
      \hat{A}_i^{(k)}
   \Bigr)
\end{aligned}
\end{equation}
where $\epsilon_c$ is the clipping parameter (distinct from the numerical stability constant~$\epsilon$ in Eq.~\ref{eq:gspo_advantage}). The clipping prevents excessively large policy updates, stabilizing training. By focusing on within-group rankings rather than absolute reward values, GSPO does not require a separate value network and reduces the sensitivity of training to the absolute scale of the reward function.

This formulation offers three properties that are relevant for structured process model generation. First, the comparison of candidates against each other rather than against an absolute baseline mitigates sensitivity to reward miscalibration, which is a relevant concern when aggregating heterogeneous quality metrics that may operate on different effective scales despite nominal normalization to~$[0,1]$. Second, the sequence-level optimization mitigates the challenge of token-wise credit assignment in long, structured outputs where individual tokens contribute to global model properties. In BPMN generation, the correctness of a gateway pair depends on tokens that may be separated by hundreds of positions; assigning token-level credit for such global properties is inherently difficult for PPO-style methods Third, group normalization stabilizes training dynamics by ensuring that policy gradients are driven by relative quality differences rather than absolute reward magnitudes, which is beneficial when the reward distribution shifts over the course of training as the model improves.

In both training stages, we employ Low-Rank Adaptation (LoRA)~\cite{hu2022lora} to maintain parameter efficiency. LoRA introduces small trainable rank-decomposition matrices into each attention layer while keeping the original model weights frozen, thereby preserving the base model's general language capabilities while allowing task-specific adaptation with a fraction of the parameters. The same base dataset is used for both SFT and GSPO to ensure that performance differences between the two stages can be attributed to the training method and reward signal rather than to data distribution shifts.

%--------------------------------------------------------------
\subsection{Process Model Quality Assessment}
\label{method:quality}

The effectiveness of RL depends critically on the reward signal. In our approach, reward signals are derived from a multi-dimensional quality assessment of the generated BPMN models based on the BEF4LLM framework~\cite{lauer2026bef4llm}, which provides an automated, metric-based evaluation grounded in established process model quality frameworks from the BPM literature, in particular the SIQ framework~\cite{reijers2010siq} and widely adopted process model complexity metrics~\cite{mendling2008metrics}. Rather than relying on a single aggregate score or a learned proxy reward model, BEF4LLM yields \emph{verifiable} component scores with regard to different model properties, that can be composed into reward signals through weighting and aggregation.

The assessment operates on the BPMN~XML representation obtained via the deterministic translator~$\phi$ and proceeds in two stages: a validity check, followed by quality evaluation across three distinct dimensions.

\paragraph{Validity.}
Given a raw output sequence $\tilde{y}$, the validity function captures the full chain of potential failures, from malformed text to trivial process models:
\begin{equation}
\label{eq:validity}
v(\tilde{y}) =
  \begin{cases}
    1 & \text{if } \tilde{y} \text{ parses to } y \in \mathcal{Y},\;
        \phi(y) \neq \bot,\;
        |FO(\phi(y))| \geq 2 \\
    0 & \text{otherwise}
  \end{cases}
\end{equation}
where $FO(\cdot)$ denotes the set of flow objects in the translated model. Validity serves as a gatekeeping criterion: only valid outputs ($v(\tilde{y}) = 1$) proceed to quality evaluation. Invalid outputs, whether due to malformed JSON, parse failures, translator errors, schema violations, or trivial structures, receive no quality scores. In the context of RL training, the validity check determines whether a candidate receives a quality-based reward or a penalty signal.

\paragraph{Syntactic Quality ($r_{\mathrm{syn}} \in [0,1]$).}
Syntactic quality measures whether the generated process model adheres to the structural rules prescribed by the BPMN modeling language. The BEF4LLM framework assesses this dimension through 16 rule-based metrics derived from the BPMN specification, as well as general applicapable modeling rules~\cite{dijkman2007bpmn,omg2013bpmn,wong2008bpmn}. These metrics span three categories of modeling rules.

\emph{Existence constraints} verify that fundamental structural elements are present: the model must contain at least one start event and one end event, and each process within the model must have exactly one of each. \emph{Degree constraints} check the connectivity of individual elements: tasks must have exactly one incoming and one outgoing sequence flow ($\mathrm{in}{=}1,\,\mathrm{out}{=}1$), start events must have no incoming and exactly one outgoing flow ($\mathrm{in}{=}0,\,\mathrm{out}{=}1$), end events the reverse ($\mathrm{in}{=}1,\,\mathrm{out}{=}0$), split gateways must fan out ($\mathrm{in}{=}1,\,\mathrm{out}{>}1$), and join gateways must converge ($\mathrm{in}{>}1,\,\mathrm{out}{=}1$). \emph{Structural constraints} address higher-level well-formedness properties, including valid sequence and message flow connections between appropriate element types, matching join gateways for every split gateway, exactly one process per pool, and descriptive labels on all observable tasks.

Each metric is computed either as a Boolean indicator (scored~0 or~1) or as the proportion of elements satisfying the rule. The syntactic quality score is the arithmetic mean of all 16~metric scores:
\begin{equation}
\label{eq:qsyn}
r_{\mathrm{syn}}
  = \frac{1}{16}\sum_{m \in \mathcal{M}_{\mathrm{syn}}} \mathrm{score}(m)
\end{equation}

As a reward signal, syntactic quality provides a direct and interpretable measure of rule conformance, capturing errors such as unmatched gateways, missing events, or improperly connected flows.

\paragraph{Pragmatic Quality ($r_{\mathrm{pra}} \in [0,1]$).}
Pragmatic quality concerns whether a process model can be understood by a human reader. Even a syntactically correct and semantically accurate process model is of limited practical value if its structure is so complex that stakeholders cannot interpret it. The BEF4LLM framework operationalizes this dimension through 15~established complexity metrics~\cite{mendling2008metrics,rolon2008bpmn,vanderfeesten2008quest} organized into five categories.

\emph{Size metrics} quantify the overall extent of the model, including the total number of nodes, gateways, sequence flows, message flows, and the diameter (longest path). \emph{Density metrics} relate the number of connections to the number of nodes, capturing how tightly interconnected the model is. \emph{Connector interplay metrics} focus on the behavior and diversity of gateways: control-flow complexity measures the routing complexity introduced by splits, while gateway heterogeneity captures the mix of gateway types (exclusive, parallel, inclusive). \emph{Partitionability metrics} assess the internal organization of the model: sequentiality measures the proportion of the process that follows a linear path, separability counts articulation points that divide the model into independent components, and depth captures the nesting level of concurrent and alternative structures. \emph{Concurrency metrics} evaluate the extent of parallel execution paths through the token split measure.

Raw metric values are mapped to normalized scores in~$[0,1]$ using a threshold-based scheme with four empirically validated thresholds per metric drawn from the literature~\cite{sanchez2011thresholds,sanchez2017bpmima,boomsma2009thresholds,ekstedt2015quality}. The thresholds partition the value range into five groups, each assigned a score from~0 to~1 in steps of~0.25, with the direction of the mapping (ascending or descending) determined by whether higher or lower values indicate better understandability. The pragmatic quality score is the arithmetic mean:
\begin{equation}
\label{eq:qpra}
r_{\mathrm{pra}}
  = \frac{1}{15}\sum_{m \in \mathcal{M}_{\mathrm{pra}}} \mathrm{score}(m)
\end{equation}

As a reward component, pragmatic quality aims towards process models that are not only correct but also concise and well-structured. This can create a tension with semantic quality, which may favor larger and more detailed models that capture every nuance of the textual description.
%, a potential trade-off that the reward weighting scheme is designed to navigate.

\paragraph{Semantic Quality ($r_{\mathrm{sem}} \in [0,1]$).}
Semantic quality evaluates the degree to which the generated model faithfully represents the target process as described in the input text. It is assessed by comparing the generated model~$M_c$ against a ground-truth reference model~$M_g$ using seven similarity metrics~\cite{dijkman2011similarity,becker2012similarity,vandongen2013similarity} that span three complementary perspectives.

\emph{Natural-language similarity} compares the textual labels of matched node pairs. Syntactic label similarity uses character-level edit distance, semantic label similarity incorporates word overlap with synonym handling, and context similarity extends the comparison to the neighborhood structure around each matched pair. \emph{Graph-structure similarity} assesses topological correspondence through graph-edit distance and through the proportion of common nodes and edges. \emph{Behavioral similarity} compares execution semantics: causal-footprint overlap captures ordering and concurrency relations, while dependency-graph overlap assesses direct succession relationships. These behavioral metrics are particularly important because two process models can be structurally different yet behaviorally equivalent, or structurally similar yet express different execution semantics.

All metrics are computed on the flow objects and connection objects of both process models. Label-based comparisons use optimal bipartite matchings to establish node correspondences, with gateway nodes excluded from label similarity computation. Each metric yields a value in~$[0,1]$, and the semantic quality score is:
\begin{equation}
\label{eq:qsem}
r_{\mathrm{sem}}
  = \frac{1}{7}\sum_{m \in \mathcal{M}_{\mathrm{sem}}} \mathrm{score}(m)
\end{equation}

Semantic quality is the most challenging dimension to optimize because it depends not only on the generated BPMN model but also on the correspondence to an external reference. It is also the dimension where LLMs have consistently shown the largest gap relative to human modelers~\cite{lauer2026bef4llm}, making it an important but challenging target for RL-driven improvement.

\paragraph{Assessment as Reward Source.}
Together, the three quality dimensions provide a comprehensive and interpretable assessment of process model quality. Each captures a distinct perspective and each is computed fully automatically from the generated BPMN model (and, for semantic quality, a ground-truth reference). The scores are bounded in~$[0,1]$ by construction, enabling consistent aggregation. Crucially, the metrics are \emph{verifiable}: unlike learned reward models that may suffer from distributional shift, each score can be traced to specific, inspectable properties of the generated model. This verifiability is a key advantage for RL training, as it ensures that improvements in reward correspond to genuine improvements in process model quality rather than exploitation of proxy weaknesses. We refer to~\cite{lauer2026bef4llm} for the complete metric definitions, normalization procedures, threshold values, and aggregation rules.

%--------------------------------------------------------------
\subsection{Reward Function Design}
\label{method:reward}

To systematically investigate how reward composition affects RL-driven model improvement, we define a family of six reward functions. Each maps the raw output~$\tilde{y}$, through its validity status and quality scores, to a scalar reward. For a valid output ($v(\tilde{y}) = 1$), let $(r_{\mathrm{syn}}, r_{\mathrm{pra}}, r_{\mathrm{sem}})$ denote the three quality scores. The reward functions share the following unified scalar formulation:
\begin{equation}
\label{eq:reward}
R(\mathbf{w}, p;\, \tilde{y})
  = \begin{cases}
      w_{\mathrm{syn}}\, r_{\mathrm{syn}}
      + w_{\mathrm{pra}}\, r_{\mathrm{pra}}
      + w_{\mathrm{sem}}\, r_{\mathrm{sem}}
        & \text{if } v(\tilde{y}) = 1 \\[6pt]
      p & \text{if } v(\tilde{y}) = 0
    \end{cases}
\end{equation}
where $\mathbf{w} = (w_{\mathrm{syn}}, w_{\mathrm{pra}}, w_{\mathrm{sem}})$ is a weight vector satisfying $\sum w_i = 1$, and $p \in \{-1, 0\}$ is the invalidity penalty. The reward for valid outputs lies in~$[0,1]$; including the penalty, the full range is~$[-1,1]$. Table~\ref{tab:reward_configs} summarizes the six configurations.

\begin{table}[t]
\centering
\caption{Reward function configurations. All weight vectors sum to~1. $R_{\mathrm{avg}}$ and~$R_1$ are mathematically equivalent but correspond to separate implementation pipelines (see text). $R_2$ isolates the effect of the penalty signal; $R_3$--$R_5$ isolate the effect of dimensional emphasis.}
\label{tab:reward_configs}
\smallskip
\resizebox{\columnwidth}{!}{
\begin{tabular}{c c c c c l}
\toprule
Config & $w_{\mathrm{syn}}$ & $w_{\mathrm{pra}}$ & $w_{\mathrm{sem}}$ & Penalty~$p$ & Emphasis \\
\midrule
$R_{\mathrm{avg}}$ & $\tfrac{1}{3}$ & $\tfrac{1}{3}$ & $\tfrac{1}{3}$ & $-1$ & Aggregated scalar \\[3pt]
$R_1$ & $\tfrac{1}{3}$ & $\tfrac{1}{3}$ & $\tfrac{1}{3}$ & $-1$ & Equal, decomposed \\[3pt]
$R_2$ & $\tfrac{1}{3}$ & $\tfrac{1}{3}$ & $\tfrac{1}{3}$ & $\phantom{-}0$ & No negative penalty \\[3pt]
$R_3$ & $\tfrac{3}{5}$ & $\tfrac{1}{5}$ & $\tfrac{1}{5}$ & $-1$ & Syntactic quality \\[3pt]
$R_4$ & $\tfrac{1}{5}$ & $\tfrac{3}{5}$ & $\tfrac{1}{5}$ & $-1$ & Pragmatic quality \\[3pt]
$R_5$ & $\tfrac{1}{5}$ & $\tfrac{1}{5}$ & $\tfrac{3}{5}$ & $-1$ & Semantic quality \\
\bottomrule
\end{tabular}
}
\end{table}

Although $R_{\mathrm{avg}}$ and~$R_1$ are mathematically equivalent in their scalar reward value (both compute the unweighted mean of the three quality scores with $p = -1$), they are retained as separate experimental runs because they correspond to two independently implemented reward pipelines. This allows us to verify whether the implementation pathway itself introduces any measurable difference. Since both configurations are equivalent in expectation, differences between them are interpreted cautiously in Section~\ref{sec:discussion}.

%--------------------------------------------------------------
\subsection{Design Rationale}
\label{method:rationale}

%Several deliberate design choices underlie the reward function family.

\paragraph{Multi-objective decomposition.}
BPMN model quality is inherently multi-dimensional: a model may be syntactically flawless yet semantically incomplete, or semantically rich yet pragmatically incomprehensible due to excessive size and structural complexity. Preserving the decomposition into interpretable components enables targeted optimization and allows observation of how RL training redistributes effort across quality dimensions.

\paragraph{Penalty versus zero reward.}
The comparison between~$R_1$ (penalty $p=-1$) and~$R_2$ (penalty $p=0$) serves as a controlled experiment on the role of negative reinforcement in structured generation. A strong penalty creates a steep gradient away from invalid outputs, providing a clear training signal to avoid producing unparseable or schema-violating output. However, when invalid outputs are frequent in the early phases of RL training, a dominant penalty signal may destabilize optimization by concentrating the loss on avoidance rather than quality improvement. Conversely, a zero penalty treats invalid outputs as uninformative rather than harmful, potentially leading to smoother optimization trajectories but weaker incentives for structural validity.

\paragraph{Weighting strategies.}
The $3{:}1{:}1$ ratio used in~$R_3$--$R_5$ represents a minimal integer weighting that introduces a clear preference for one quality objective while maintaining non-trivial contributions from the other two dimensions. The weights are not intended to be optimal in any absolute sense; rather, they provide a controlled mechanism for analyzing the sensitivity of LLM performance to different reward priorities. More aggressive ratios (e.g., $5{:}1{:}1$) risk collapsing the multi-objective signal into near-single-objective optimization, while more moderate ratios (e.g., $2{:}1{:}1$) may produce insufficient differentiation. The $3{:}1{:}1$ scheme balances experimental discriminability with training stability.

\paragraph{Shared training data.}
Using the same underlying dataset for both SFT and GSPO ensures that performance differences between the two stages, and across reward configurations, can be attributed to the training method and reward signal rather than to differences in data distribution or domain coverage.

\paragraph{Bounded and normalized rewards.}
Since all constituent metrics are bounded in~$[0,1]$ by the evaluation framework, the composite reward for valid outputs is likewise bounded in~$[0,1]$, and the full reward range including penalties lies in~$[-1,1]$. This ensures consistent scaling across reward configurations and avoids the need for additional reward clipping beyond the group-wise normalization already performed by GSPO.

%--------------------------------------------------------------
\subsection{Grammar-Constrained Decoding}
\label{method:grammar}

As an additional experimental factor orthogonal to the reward function design, we investigate the effect of grammar-constrained decoding during inference. When enabled, the decoding process restricts the LLM's token-level sampling to outputs conforming to a predefined Extended Backus–Naur form (EBNF) derived from the expected output schema. Concretely, at each generation step, the set of permissible next tokens is filtered to include only those that would maintain conformance with the grammar, thereby guaranteeing that every completed output is at minimum a structurally well-formed JSON document with the expected top-level fields.
It is important to note, however, that even with EBNF active, the LLM can still produce outputs that are invalid, e.g., in the context of BPMN rules a cross-pool sequence flow.
 
Grammar-constrained decoding does not alter the learned policy: it is applied exclusively at inference time as a structural guard. Its inclusion in the experimental design allows us to disentangle two fundamentally different sources of structural correctness (correctness \emph{learned} through reward-driven optimization versus correctness \emph{enforced} through external decoding constraints) and to assess whether these mechanisms are complementary, additive, or largely redundant.
\section{Experimental Setup}
\label{sec:experiment}

%This section describes the experimental configuration used to evaluate the proposed RL-based training pipeline. We present the dataset (Section~\ref{sec:dataset}), the selected LLMs (Section~\ref{sec:models}), the training setup for both pipeline stages (Section~\ref{sec:training}), the evaluation protocol (Section~\ref{sec:eval_protocol}), the statistical methodology (Section~\ref{sec:stat_method}), and the research questions (Section~\ref{sec:rqs}).

%--------------------------------------------------------------
\subsection{Dataset}
\label{sec:dataset}

A common data source is used to construct both the training and held-out evaluation sets, with no overlap between them. The same training set is used for all SFT and GSPO runs, and the same held-out evaluation set is used for all configurations. This design reduces the risk that observed performance differences are driven by data distribution differences rather than by the training method, reward signal, or initialization strategy.

\paragraph{Training set.}
The training corpus comprises 1,552 samples, each consisting of four components: a system prompt defining the modeling task and the expected output format, a concise modeling instruction, a textual process description serving as input, and the corresponding target process model. The textual descriptions are a mix of German and English, reflecting the bilingual nature of the source datasets and the practical requirement that process modeling tools operate across language boundaries. The training data was assembled from multiple publicly available BPMN datasets, supplemented with synthetically generated samples to increase corpus size and diversity. The same training set is used for both SFT and GSPO to ensure comparability between the two training stages and across reward configurations.

\paragraph{Evaluation set.}
A held-out evaluation set of 105 text-process model pairs is used throughout all experiments. Each pair consists of a textual process description and a corresponding ground-truth BPMN model, the latter being required for computing semantic quality scores. The evaluation samples are drawn from publicly available datasets, including the Camunda BPMN for Research collection and the dataset by~\citet{mangler2023dataset}, and have been manually verified for quality. No evaluation sample appears in the training set. All ground-truth BPMN models have more than two flow objects, ensuring that the evaluation covers non-trivial process structures.

%--------------------------------------------------------------
\subsection{LLM Selection}
\label{sec:models}

The experiments cover two LLM families: Llama~3.1 8B~Instruct and Qwen~2.5 14B~Instruct. These were selected to provide diversity along two axes: parameter count (8B vs.\ 14B) and architecture family (Llama vs.\ Qwen), enabling analysis of whether findings generalize across model sizes and pre-training regimes. Both LLMs are instruction-tuned variants, which are more likely to follow the structured output requirements of the task.
The choice to focus on small-to-medium-sized LLMs rather than the largest available LLMs is deliberate. These LLMs represent the practical deployment frontier for organizations that require local inference without reliance on commercial APIs, a relevant consideration for enterprise process modeling where confidentiality constraints often preclude sending process descriptions to external services. Prior work~\cite{lauer2026bef4llm} showed that models in this size range already achieve competitive BPMN modeling quality, making them suitable candidates for targeted improvement through reinforcement learning.
The selected LLMs are well-established baselines with broad adoption in SFT and RLVR research. More recent releases in these families are often optimized for agentic and tool-use scenarios including multi-step reasoning. We argue that these capabilities do not necessarily translate to a meaningful advantage for producing BPMN models from natural language process descriptions.

%--------------------------------------------------------------
\subsection{Training Configuration}
\label{sec:training}

Each training stage, when applied, employs Low-Rank Adaptation (LoRA)~\cite{hu2022lora} and operates in bfloat16 precision to ensure numerical stability.

\paragraph{Stage~1: SFT}
Supervised fine-tuning is performed using the LLaMA-Factory framework\footnote{\url{https://github.com/hiyouga/LlamaFactory}}. LoRA is applied with rank~8 targeting all compatible modules, a dropout of~0.1, and a learning rate of $1 \times 10^{-4}$ with cosine scheduling and a warmup ratio of~0.1. Training runs for 3~epochs with a per-device batch size of~1 and 8~gradient accumulation steps, yielding an effective batch size of~8. The loss is computed solely on the output tokens. Table~\ref{tab:sft_params} summarizes the SFT hyperparameters.

\begin{table}[t]
\centering
\caption{SFT hyperparameter settings. Remaining parameters are kept at LLaMA-Factory defaults.}
\label{tab:sft_params}
\smallskip
\begin{tabular}{l l}
\toprule
Parameter & Value \\
\midrule
LoRA rank & 8 \\
LoRA target modules & all \\
LoRA dropout & 0.1 \\
Learning rate & $1 \times 10^{-4}$ \\
Warmup ratio & 0.1 \\
LR scheduler & cosine \\
Per-device batch size & 1 \\
Gradient accumulation steps & 8 \\
Epochs & 3 \\
Precision & bfloat16 \\
\bottomrule
\end{tabular}
\end{table}

\paragraph{Stage~2: GSPO}
GSPO is performed using the unsloth framework\footnote{\url{https://unsloth.ai/}}. LoRA rank is increased to~16 to provide additional capacity for policy adaptation, while all other LoRA settings remain consistent. The learning rate is reduced to $5 \times 10^{-5}$, and weight decay of~0.1 is applied to regularize updates. For candidate generation, $K = 4$ outputs are sampled per prompt using a temperature of~0.7 and top-$p$ of~0.95 to encourage diverse yet coherent candidates within each group. Training runs for 1~epoch with a per-device batch size of~1 and 8~gradient accumulation steps. The optimizer is paged AdamW with 8-bit quantization to reduce memory overhead. Table~\ref{tab:gspo_params} summarizes the GSPO hyperparameters.

\begin{table}[t]
\centering
\caption{GSPO hyperparameter settings. Remaining parameters are kept at Unsloth defaults.}
\label{tab:gspo_params}
\smallskip
\begin{tabular}{l l}
\toprule
Parameter & Value \\
\midrule
LoRA rank & 16 \\
LoRA target modules & all \\
Learning rate & $5 \times 10^{-5}$ \\
Weight decay & 0.1 \\
Warmup ratio & 0.1 \\
Max gradient norm & 1.0 \\
Temperature & 0.7 \\
Top-$p$ & 0.95 \\
LR scheduler & cosine \\
Optimizer & paged AdamW 8-bit \\
Per-device batch size & 1 \\
Gradient accumulation steps & 8 \\
Number of generations ($K$) & 4 \\
Epochs & 1 \\
Precision & bfloat16 \\
\bottomrule
\end{tabular}
\end{table}

%Several deliberate differences between the SFT and GSPO configurations merit explanation. 
The increased LoRA rank ($8 \rightarrow 16$) provides additional representational capacity for the policy to learn quality-dependent adjustments beyond what imitation learning established. The reduced learning rate ($10^{-4} \rightarrow 5 \times 10^{-5}$) reflects the need for more conservative updates during RL, where overly aggressive optimization can destabilize the policy. The single-epoch GSPO training is intended to reduce the risk of reward overfitting, a known concern in RL for language models where extended training can lead the model to exploit reward function artifacts rather than genuinely improving output quality. The sampling parameters (temperature~0.7, top-$p$~0.95) are chosen to ensure sufficient diversity within each candidate group, as GSPO's group-relative advantage computation requires meaningful variation to produce informative gradients.

\paragraph{Experimental matrix.}
Throughout this section, \emph{untrained} refers to the task-specific condition in which GSPO is applied directly to the original instruction-tuned base model without prior SFT. It does not denote a randomly initialized or generally untrained LLM. For each of the two LLMs, we evaluate all combinations of:
\begin{itemize}
    \item \emph{Base model}: untrained (i.e., instruction-tuned base model available on HuggingFace\footnote{\url{https://huggingface.co/}}) and SFT-initialized.
    \item \emph{Reward function (Section~\ref{method:reward})}: $R_{\mathrm{avg}}$, $R_1$, $R_2$, $R_3$, $R_4$, $R_5$.
    \item \emph{Grammar-constrained decoding (Section~\ref{method:grammar})}: enabled and disabled.
\end{itemize}
This yields $2 \times 6 \times 2 = 24$ configurations per LLM, which, across two LLMs, results in a total of 48 RL-trained configurations. In addition to these RL-trained configurations, we evaluate the SFT-only model for each LLM as a non-RL baseline. Section~\ref{sec:results} reports results for the no-grammar condition to isolate the effects of reward design and initialization; the interaction with grammar-constrained decoding is analyzed in Section~\ref{sec:discussion}.

%--------------------------------------------------------------
\subsection{Evaluation Protocol}
\label{sec:eval_protocol}

Each trained model is evaluated on the full set of 105 held-out samples. For each sample, the model generates a raw output sequence~$\tilde{y}$, which undergoes the parsing, translation, and validation chain defined in Section~\ref{method:task}. The evaluation proceeds in two stages:
\begin{enumerate}
    \item \textbf{Validity check.} The raw output is assessed via the validity function $v(\tilde{y})$ (Eq.~\ref{eq:validity}): parsing into an intermediate model, translation via~$\phi$, BPMN~2.0 schema conformance, and a minimum of two flow objects must all succeed. Only valid outputs ($v(\tilde{y}) = 1$) proceed to quality evaluation.
    \item \textbf{Quality scoring.} For valid outputs, the three dimension scores $r_{\mathrm{syn}}$, $r_{\mathrm{pra}}$, and $r_{\mathrm{sem}}$ are computed on the translated BPMN~XML. Invalid outputs receive no quality scores; their contribution is captured solely through the validity count.
\end{enumerate}

Reported quality scores per configuration are the arithmetic mean over all valid samples. This means that the syntactic, pragmatic, and semantic scores are conditioned on validity: they describe the quality of the models the LLM \emph{successfully produces}, not the quality of all attempted outputs. This conditioning is necessary because quality metrics cannot be computed for invalid BPMN~XML, and it is consistent with the evaluation protocol established in~\cite{lauer2026bef4llm}.

Each configuration generates a single output per evaluation sample. Unlike the multi-run protocol in~\cite{lauer2026bef4llm}, we do not perform repeated generation runs, which limits the ability to estimate within-configuration variance but keeps the total computational cost tractable given the 48 configurations evaluated.

%--------------------------------------------------------------
\subsection{Statistical Methodology}
\label{sec:stat_method}

To determine whether observed performance differences are statistically significant, we adopt a non-parametric permutation testing framework~\cite{good2005permutation} appropriate for paired data with no distributional assumptions. For each pair of configurations to be compared (e.g., $R_1$ vs.\ $R_3$ under the same base model and grammar setting), we test the null hypothesis $H_0\!: \mathbb{E}[a_i - b_i] = 0$, where $a_i$ and $b_i$ are the quality scores of configurations~$A$ and~$B$ on evaluation sample~$i$, against the two-sided alternative. The test is conducted on paired samples; only those evaluation instances for which both configurations produced a valid BPMN model are included. The test statistic is the mean paired difference $\bar{d} = \frac{1}{n} \sum_{i=1}^{n} (a_i - b_i)$. Under~$H_0$, the sign of each paired difference is equally likely to be positive or negative; significance is assessed by randomly flipping the signs of the observed differences across 99{,}999 permutations and computing the proportion of permuted statistics whose absolute value equals or exceeds~$|\bar{d}|$. The number of resamples ensures a minimum attainable $p$-value of $2 \times 10^{-5}$, well below the significance threshold after Bonferroni correction. Unlike the Wilcoxon signed-rank test, the permutation test makes no assumptions about the symmetry of the difference distribution, making it robust to the skewed and bounded score distributions (values in $[0,1]$ with ceiling effects) that arise in our setting.

Because quality metrics can only be computed for valid BPMN outputs, all comparisons are conditioned on joint validity; a sample contributes to the test only if both configurations produce a valid model for that sample. This complete-case design controls for sample difficulty but implies that the results describe quality differences among successfully generated BPMN models, not among all generation attempts. Configurations with substantially different validity rates (e.g., SFT-only with 36 valid samples vs.\ $R_1$ with 102) are compared on a restricted subset of samples, which should be considered when interpreting effect sizes. Validity counts are therefore reported alongside all quality scores to ensure that improvements in conditional quality are not interpreted independently of the model's ability to produce valid outputs.

Tests are conducted independently for syntactic, pragmatic, and semantic quality, as the three dimensions capture qualitatively distinct aspects that may respond differently to the same experimental manipulation. Where multiple pairwise tests are conducted within the same analysis, raw $p$-values are adjusted using Bonferroni correction to control the family-wise error rate at $\alpha = 0.05$. Specifically, for RQ1 (one comparison across three dimensions) and for the penalty and initialization analyses (RQ2b, RQ3), the correction factor is $m{=}3$; for the reward weighting analysis (RQ2a: three reward comparisons $\times$ three dimensions), $m{=}9$. A difference is deemed significant only if the adjusted $p$-value falls below~0.05. Throughout, we denote significance as \textsuperscript{***} ($p < 0.001$), \textsuperscript{**} ($p < 0.01$), \textsuperscript{*} ($p < 0.05$), or n.s.\ (not significant). Beyond statistical significance, we report the mean paired difference~$\Delta$ to assess practical magnitude. Small but statistically significant differences (e.g., $\Delta < 0.01$) are noted as such and interpreted with appropriate caution, recognizing that statistical significance does not automatically imply practical relevance.

%--------------------------------------------------------------
\subsection{Research Questions}
\label{sec:rqs}

Given the experimental design described above, we investigate three research questions:

\medskip
\noindent\textbf{RQ1:} \emph{Does GSPO with domain-grounded reward signals improve process model quality beyond SFT?}

This question establishes the basic value of the RL approach. We compare representative GSPO configurations, primarily the equal-weight reference configuration~$R_1$, against SFT-only baselines across all three quality dimensions, assessing both the magnitude and statistical significance of the improvement.

\medskip
\noindent\textbf{RQ2:} \emph{How does reward function composition, specifically dimensional weighting and the invalidity penalty, affect the balance across quality dimensions?}

This question addresses the central design challenge: given a multi-dimensional quality framework, can the reward function be used to steer the optimization toward desired quality profiles? We investigate two compositional choices (the relative weighting of quality dimensions, $R_1$ vs.\ $R_3$--$R_5$, and the penalty for invalid outputs, $R_1$ vs.\ $R_2$) and analyze their interaction with model architecture.

\medskip
\noindent\textbf{RQ3:} \emph{How does the choice of base model initialization (SFT-initialized versus untrained) affect RL optimization outcomes?}

This question examines the standard two-stage pipeline assumption. By training GSPO from both SFT-initialized and untrained bases under identical reward configurations, we assess whether SFT is a necessary prerequisite for effective RL in the context of process modeling or whether it can be bypassed, and whether the answer depends on the model architecture.
\section{Results}
\label{sec:results}

%This section presents the empirical findings, organized around three research questions. Section~\ref{sec:rq1} examines whether GSPO improves process model quality beyond SFT. Section~\ref{sec:rq2} investigates how reward function composition shapes quality trade-offs. Section~\ref{sec:rq3} analyzes the role of base model initialization. The analysis focuses on Llama~3.1 8B~Instruct and Qwen~2.5 14B~Instruct, the two model families for which complete per-sample data is available.

Table~\ref{tab:main_results} provides the complete descriptive results for all experimental configurations, reporting the mean quality score per dimension computed over valid outputs alongside the number of valid samples (out of 105 evaluation instances).

% ============================================================
% MAIN RESULTS TABLE
% ============================================================
\begin{table*}[h]
\centering
\caption{Quality scores across all experimental configurations (no grammar). Scores are the arithmetic mean over valid samples. $n$ denotes the number of valid outputs out of 105 evaluation samples. Bold indicates the highest conditional quality score per column within each model--base combination; scores should be interpreted together with the corresponding validity count~$n$. \emph{Untrained} denotes the original instruction-tuned base model without BPM-specific SFT (see Section~\ref{sec:training}).}
\label{tab:main_results}
\smallskip
\resizebox{\textwidth}{!}{%
\begin{tabular}{ll l rrr r rrr r}
\toprule
& & & \multicolumn{4}{c}{\textbf{Llama 3.1 8B}} & \multicolumn{4}{c}{\textbf{Qwen 2.5 14B}} \\
\cmidrule(lr){4-7} \cmidrule(lr){8-11}
Base & Reward & Emphasis & $r_{\mathrm{syn}}$ & $r_{\mathrm{pra}}$ & $r_{\mathrm{sem}}$ & $n$ & $r_{\mathrm{syn}}$ & $r_{\mathrm{pra}}$ & $r_{\mathrm{sem}}$ & $n$ \\
\midrule
\multicolumn{2}{l}{\emph{SFT-only (no RL)}} & baseline & 0.824 & 0.794 & 0.561 & 36 & 0.869 & 0.807 & 0.595 & 78 \\
\midrule
SFT & $R_{\mathrm{avg}}$ & aggregated scalar & 0.869 & 0.923 & 0.592 & 101 & 0.876 & 0.922 & \textbf{0.584} & 101 \\
    & $R_1$ & equal, $p{=}{-1}$ & \textbf{0.926} & \textbf{0.934} & \textbf{0.594} & 102 & \textbf{0.888} & 0.923 & 0.567 & 101 \\
    & $R_2$ & equal, $p{=}0$ & 0.917 & 0.931 & \textbf{0.594} & 99 & 0.750 & \textbf{0.964} & 0.547 & 104 \\
    & $R_3$ & syntax $\uparrow$ & 0.662 & 0.884 & 0.551 & 82 & 0.881 & 0.930 & 0.585 & 99 \\
    & $R_4$ & pragmatic $\uparrow$ & 0.871 & 0.813 & 0.593 & 82 & 0.812 & 0.946 & 0.592 & 102 \\
    & $R_5$ & semantic $\uparrow$ & 0.801 & 0.822 & 0.564 & 38 & 0.816 & 0.962 & 0.548 & 92 \\
\midrule
Untrained & $R_{\mathrm{avg}}$ & aggregated scalar & 0.839 & \textbf{0.937} & 0.577 & 39 & 0.895 & 0.904 & \textbf{0.609} & 87 \\
          & $R_1$ & equal, $p{=}{-1}$ & 0.664 & 0.885 & 0.546 & 88 & 0.875 & 0.920 & 0.601 & 94 \\
          & $R_2$ & equal, $p{=}0$ & 0.850 & 0.923 & 0.583 & 76 & 0.784 & 0.935 & 0.598 & 97 \\
          & $R_3$ & syntax $\uparrow$ & 0.769 & 0.843 & 0.540 & 26 & \textbf{0.898} & 0.920 & 0.586 & 90 \\
          & $R_4$ & pragmatic $\uparrow$ & 0.836 & 0.929 & 0.517 & 83 & 0.869 & \textbf{0.938} & 0.584 & 88 \\
          & $R_5$ & semantic $\uparrow$ & \textbf{0.867} & 0.935 & \textbf{0.588} & 93 & 0.928 & 0.908 & 0.594 & 96 \\
\bottomrule
\end{tabular}%
}
\end{table*}

% ============================================================
% CONSOLIDATED STATISTICAL TESTS TABLE
% ============================================================

Table~\ref{tab:stat_tests} consolidates all pairwise statistical tests referenced throughout this section.

\begin{table*}[]
\centering
\caption{Consolidated pairwise permutation tests (Bonferroni-adjusted, 99{,}999 resamples). All comparisons use the no-grammar condition. $\Delta$ is the mean paired difference $A - B$. Significance: \textsuperscript{***}$p < 0.001$, \textsuperscript{**}$p < 0.01$, \textsuperscript{*}$p < 0.05$, n.s.\ = not significant. ``Unt'' denotes the untrained (BPM-unadapted instruction-tuned) base.}
\label{tab:stat_tests}
\smallskip
\resizebox{\textwidth}{!}{%
\begin{tabular}{ll ll c rrr rrr rrr}
\toprule
& & & & & \multicolumn{3}{c}{\textbf{Syntactic}} & \multicolumn{3}{c}{\textbf{Pragmatic}} & \multicolumn{3}{c}{\textbf{Semantic}} \\
\cmidrule(lr){6-8} \cmidrule(lr){9-11} \cmidrule(lr){12-14}
RQ & Model & $A$ & $B$ & $m$ & $\Delta$ & $p_{\mathrm{adj}}$ & & $\Delta$ & $p_{\mathrm{adj}}$ & & $\Delta$ & $p_{\mathrm{adj}}$ & \\
\midrule
\multirow{2}{*}{1} & Llama & SFT+$R_1$ & SFT-only & 3 & $+$0.092 & $6.0 \times 10^{-5}$ & \textsuperscript{***} & $+$0.139 & $6.0 \times 10^{-5}$ & \textsuperscript{***} & $+$0.030 & $1.8 \times 10^{-2}$ & \textsuperscript{*} \\
 & Qwen & SFT+$R_1$ & SFT-only & 3 & $+$0.032 & $3.4 \times 10^{-2}$ & \textsuperscript{*} & $+$0.116 & $6.0 \times 10^{-5}$ & \textsuperscript{***} & $-$0.018 & $3.5 \times 10^{-1}$ & n.s. \\
\midrule
\multirow{6}{*}{2a} & Llama & $R_1$ & $R_3$ (syn$\uparrow$) & 9 & $+$0.264 & $1.8 \times 10^{-4}$ & \textsuperscript{***} & $+$0.050 & $1.8 \times 10^{-4}$ & \textsuperscript{***} & $+$0.035 & $1.8 \times 10^{-4}$ & \textsuperscript{***} \\
 & Llama & $R_1$ & $R_4$ (pra$\uparrow$) & 9 & $+$0.052 & $3.6 \times 10^{-4}$ & \textsuperscript{***} & $+$0.120 & $1.8 \times 10^{-4}$ & \textsuperscript{***} & $-$0.001 & $1.00$ & n.s. \\
 & Llama & $R_1$ & $R_5$ (sem$\uparrow$) & 9 & $+$0.128 & $1.8 \times 10^{-4}$ & \textsuperscript{***} & $+$0.111 & $1.8 \times 10^{-4}$ & \textsuperscript{***} & $+$0.027 & $1.1 \times 10^{-3}$ & \textsuperscript{**} \\
 & Qwen & $R_1$ & $R_3$ (syn$\uparrow$) & 9 & $+$0.001 & $1.00$ & n.s. & $-$0.008 & $1.9 \times 10^{-2}$ & \textsuperscript{*} & $-$0.013 & $4.1 \times 10^{-1}$ & n.s. \\
 & Qwen & $R_1$ & $R_4$ (pra$\uparrow$) & 9 & $+$0.079 & $1.8 \times 10^{-4}$ & \textsuperscript{***} & $-$0.023 & $1.8 \times 10^{-4}$ & \textsuperscript{***} & $-$0.022 & $1.3 \times 10^{-2}$ & \textsuperscript{*} \\
 & Qwen & $R_1$ & $R_5$ (sem$\uparrow$) & 9 & $+$0.071 & $1.8 \times 10^{-4}$ & \textsuperscript{***} & $-$0.038 & $1.8 \times 10^{-4}$ & \textsuperscript{***} & $+$0.019 & $1.7 \times 10^{-2}$ & \textsuperscript{*} \\
\midrule
\multirow{2}{*}{2b} & Llama & $R_1$\,($p{=}{-1}$) & $R_2$\,($p{=}0$) & 3 & $+$0.011 & $3.5 \times 10^{-1}$ & n.s. & $+$0.002 & $1.00$ & n.s. & $-$0.001 & $1.00$ & n.s. \\
 & Qwen & $R_1$\,($p{=}{-1}$) & $R_2$\,($p{=}0$) & 3 & $+$0.138 & $6.0 \times 10^{-5}$ & \textsuperscript{***} & $-$0.041 & $6.0 \times 10^{-5}$ & \textsuperscript{***} & $+$0.021 & $2.5 \times 10^{-3}$ & \textsuperscript{**} \\
\midrule
\multirow{2}{*}{3} & Llama & SFT+$R_1$ & Unt+$R_1$ & 3 & $+$0.262 & $6.0 \times 10^{-5}$ & \textsuperscript{***} & $+$0.049 & $6.0 \times 10^{-5}$ & \textsuperscript{***} & $+$0.036 & $6.0 \times 10^{-5}$ & \textsuperscript{***} \\
 & Qwen & SFT+$R_1$ & Unt+$R_1$ & 3 & $+$0.017 & $1.6 \times 10^{-1}$ & n.s. & $+$0.003 & $7.1 \times 10^{-1}$ & n.s. & $-$0.032 & $6.0 \times 10^{-5}$ & \textsuperscript{***} \\
\bottomrule
\end{tabular}%
}
\end{table*}

% ============================================================
\subsection{RQ1: Does GSPO Improve Process Model Quality Beyond SFT?}
\label{sec:rq1}

To assess whether RL with domain-grounded rewards yields measurable quality improvements over SFT alone, we compare GSPO configurations against the SFT-only baseline. For each model, we focus on~$R_1$, the equal-weight reference configuration ($p{=}{-1}$), and additionally report~$R_{\mathrm{avg}}$ and~$R_2$ to contextualize the findings. Statistical comparisons use paired permutation tests with Bonferroni correction ($m{=}3$).

\subsubsection{Llama3.1 8B}
\label{results_llama}

GSPO with~$R_1$ produces statistically significant improvements across all three quality dimensions. Syntactic quality increases from 0.824 to 0.926 ($\Delta{=}+0.092$, $p_{\mathrm{adj}}{<}10^{-4}$), pragmatic quality from 0.794 to 0.934 ($\Delta{=}+0.139$, $p_{\mathrm{adj}}{<}10^{-4}$), and semantic quality from 0.561 to 0.594 ($\Delta{=}+0.030$, $p_{\mathrm{adj}}{=}0.018$). The improvements are not uniform across dimensions: pragmatic quality shows the largest gain, followed by syntactic, with semantic showing the smallest. This pattern recurs throughout the experiments and appears to reflect an inherent ordering in how accessible each dimension is to reward-based optimization. In practical terms, the syntactic gain corresponds to approximately 1.5 additional BPMN rules satisfied per model (out of~16), reducing average rule violations from~2.8 to~1.2. The pragmatic gain translates to roughly 2.1 additional complexity metrics falling within empirically validated acceptability thresholds (out of~15), bringing nearly all pragmatic indicators into the ``understandable'' range.

Because the SFT-only baseline produces only 36 valid outputs (out of~105), while~$R_1$ produces~102, the paired statistical comparison is conducted on a restricted subset of samples. The quality scores for the SFT-only baseline should therefore be interpreted in conjunction with its substantially lower validity rate.

Beyond the shift in mean scores, GSPO produces a qualitative change in the \emph{consistency} of generated outputs. The pragmatic quality standard deviation decreases from 0.085 under SFT-only to 0.013 under~$R_1$, a reduction by a factor of approximately~6.5. This is visible in Figure~\ref{fig:rq1} (top row): the SFT-only violins exhibit long lower tails, particularly for syntactic and pragmatic quality, indicating that SFT produces a heterogeneous mix of well-structured and poorly structured outputs depending on the input. Under GSPO, these tails are largely eliminated and the distributions compress into tight high-quality clusters. The semantic quality violins, by contrast, show substantial overlap between conditions, consistent with the smaller effect size and suggesting that semantic quality is inherently more variable and harder to stabilize through reward-based optimization.

This consistency finding is practically significant: in deployment, a model that reliably produces good process models is more useful than one that alternates between excellent and poor outputs. This suggests that GSPO's group-relative comparison mechanism implicitly penalizes high-variance generation strategies, pushing the model toward more deterministic, reliable behavior.

\subsubsection{Qwen2.5 14B}

GSPO with~$R_1$ significantly improves both pragmatic and syntactic quality for Qwen2.5, while leaving semantic quality mostly unaffected. Pragmatic quality increases from 0.807 to 0.923 ($\Delta{=}+0.116$, $p_{\mathrm{adj}}{<}10^{-4}$), and syntactic quality shows a modest but significant gain ($\Delta{=}+0.032$, $p_{\mathrm{adj}}{=}0.034$). Semantic quality shows a small numerical decrease from 0.595 to 0.567 ($\Delta{=}-0.018$), but this difference does not reach statistical significance ($p_{\mathrm{adj}}{=}0.35$), indicating that GSPO optimization does not come at a measurable cost to semantic fidelity in this case.

\begin{figure*}[t]
    \centering
    \includegraphics[width=\textwidth]{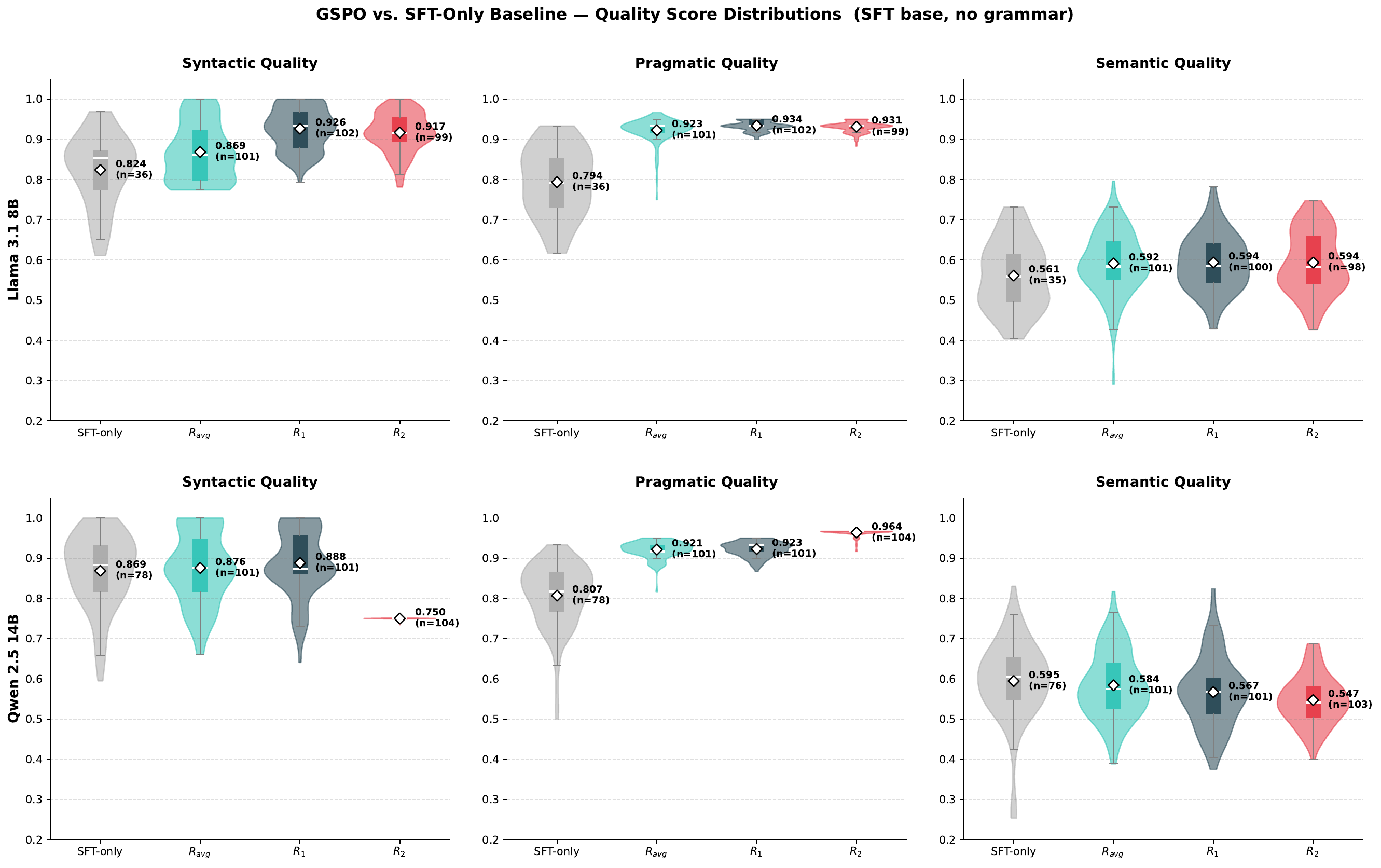}
    \caption{RQ1: Distributional comparison of quality scores between SFT-only baseline and GSPO configurations ($R_{\mathrm{avg}}$, $R_1$, $R_2$). Top row: Llama~3.1 8B. Bottom row: Qwen~2.5 14B. Violins show the kernel density estimate; embedded box plots indicate median and interquartile range; diamond markers denote the mean. All GSPO configurations use the SFT base without grammar enforcement.}
    \label{fig:rq1}
\end{figure*}

Figure~\ref{fig:rq1} (bottom row) illustrates the distributional patterns. The pragmatic quality violin for~$R_1$ is dramatically tighter and higher than the SFT-only baseline, mirroring previous observations in section \ref{results_llama} and confirming that pragmatic improvement through output regularization is a consistent effect of GSPO across model architectures. The syntactic improvement, while statistically significant, is smaller than for Llama, reflecting Qwen's higher SFT-only baseline (0.869 vs.\ 0.824), which leaves less room for structural improvement. The semantic quality violins overlap substantially between conditions, consistent with the non-significant test result and reinforcing the observation that semantic quality is the dimension least amenable to reward-based optimization.

That Qwen's semantic quality does not significantly change under GSPO, despite the model generating structurally different outputs, suggests that the equal-weight reward function achieves a balanced optimization that avoids trading semantic fidelity for structural improvements. This contrasts with configurations that push more aggressively toward pragmatic quality (e.g., $R_2$ with no penalty, which achieves 0.964 pragmatic but drops to 0.547 semantic), as analyzed under RQ2.

\subsubsection{Summary}

GSPO with equal-weight rewards consistently and significantly improves pragmatic quality for both model families, with large effect sizes ($\Delta > 0.11$, $p < 10^{-4}$). Syntactic quality improves for both models: strongly for Llama ($\Delta{=}+0.092$, $p < 10^{-4}$), more modestly for Qwen ($\Delta{=}+0.032$, $p{=}0.034$). Semantic quality shows a small positive effect for Llama ($p{=}0.018$) and no significant change for Qwen ($p{=}0.35$), indicating that the equal-weight reward achieves structural and comprehensibility improvements without sacrificing content fidelity. The most consistent effect of GSPO is the distributional tightening: RL-trained models produce substantially more consistent outputs, with pragmatic quality variability reduced by more than a factor of six.

% ============================================================
\subsection{RQ2: How Does Reward Function Composition Affect Quality Trade-Offs?}
\label{sec:rq2}

This subsection examines two aspects of reward function design: the effect of dimensional weighting (Section~\ref{sec:rq2_weights}) and the role of the invalidity penalty (Section~\ref{sec:rq2_penalty}). Together, these analyses investigate whether the reward function can serve as a steering mechanism to prioritize specific quality dimensions, and how structural design choices shape the optimization trajectory.

\subsubsection{Effect of Dimensional Weighting}
\label{sec:rq2_weights}

We compare the equal-weight reference configuration~$R_1$ against the three weighted variants~$R_3$ (syntax-emphasized, $3{:}1{:}1$), $R_4$ (pragmatic-emphasized, $1{:}3{:}1$), and~$R_5$ (semantic-emphasized, $1{:}1{:}3$), all using the SFT base. Pairwise permutation tests are Bonferroni-corrected with $m{=}9$ (three comparisons $\times$ three dimensions). Full results are reported in Table~\ref{tab:stat_tests}.

\paragraph{Llama3.1 8B.}
%The results for Llama are unequivocal and counter-intuitive:
$R_1$ statistically dominates all three weighted variants across \emph{all} quality dimensions, not just the de-emphasized ones. The most striking finding concerns~$R_3$ (syntax-weighted): despite receiving 60\% of the reward weight for syntactic quality, this configuration achieves a syntactic score of only 0.662, a drop of 0.264 points from~$R_1$'s 0.926 ($p_{\mathrm{adj}}{<}10^{-3}$). The model undergoes what we term \emph{policy collapse} (convergence to a narrow, low-diversity output mode), shifting from producing mostly rule-conformant models to producing mostly rule-violating ones. The pattern repeats for the other weighted variants: $R_4$ (pragmatic-weighted) produces \emph{lower} pragmatic quality than~$R_1$ ($\Delta{=}+0.120$, $p_{\mathrm{adj}}{<}10^{-3}$, favoring~$R_1$), and $R_5$ (semantic-weighted) achieves significantly \emph{lower} semantic quality than~$R_1$ ($\Delta{=}+0.027$, $p_{\mathrm{adj}}{=}0.001$), the very dimension it was designed to optimize, while also depressing syntactic quality by 0.128 points. Because~$R_5$ produces only 38 valid outputs, this comparison should be interpreted with the low validity count in mind.

\begin{figure*}[t]
    \centering
    \includegraphics[width=\textwidth]{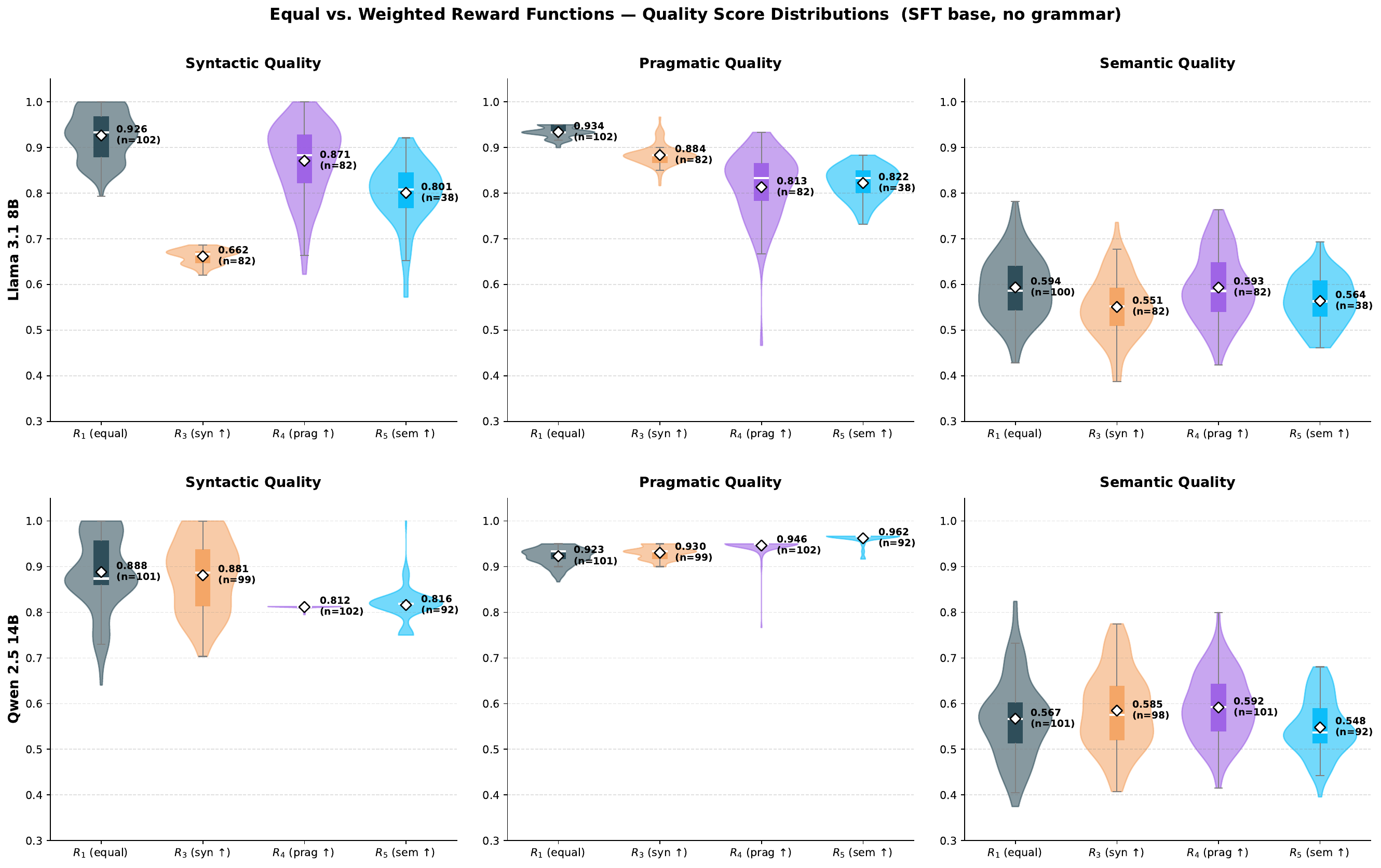}
    \caption{RQ2: Distributional comparison of quality scores across reward weighting schemes. Top row: Llama~3.1 8B. Bottom row: Qwen~2.5 14B. $R_1$ (equal) maintains broad, high distributions, while weighted variants ($R_3$--$R_5$) produce narrower distributions shifted toward lower values, most dramatically for Llama's~$R_3$, which collapses to a tight cluster at 0.662 despite emphasizing syntactic quality.}
    \label{fig:rq2_weights}
\end{figure*}

Figure~\ref{fig:rq2_weights} (top row) reveals the distributional character of these failures. The~$R_1$ violins are consistently the broadest at the top of the quality range, indicating robust high-quality generation across diverse process descriptions. The~$R_3$ syntactic violin, by contrast, forms a tight low cluster at 0.662; the model has converged to a narrow attractor from which it produces structurally similar, low-quality outputs regardless of input.~$R_4$ and~$R_5$ show intermediate patterns: wider distributions but shifted downward, suggesting that the weighted reward pushes the policy into suboptimal regions without achieving the tight, high-quality mode that~$R_1$ finds.

The failure of weighted rewards to improve their target dimension contradicts the naive expectation that more weight implies better performance along the weighted dimension. Instead, the $3{:}1{:}1$ weighting appears to destabilize the optimization by reducing the effective information content of the reward signal. When one component dominates, the relative ranking within candidate groups is driven primarily by variation on that single dimension, which may be insufficient to provide stable, informative gradients, particularly when the SFT-initialized policy already produces similar outputs along the dominant dimension..

\paragraph{Qwen 2.5 14B.}
The Qwen results are more nuanced, showing that weighted rewards can partially achieve their intended steering effect, but always with collateral consequences. $R_4$ (pragmatic-weighted) successfully improves pragmatic quality over~$R_1$ ($\Delta{=}+0.023$, $p_{\mathrm{adj}}{<}10^{-3}$) and also improves semantic quality ($\Delta{=}+0.022$, $p_{\mathrm{adj}}{=}0.013$), but at significant cost to syntactic quality ($\Delta{=}-0.079$, $p_{\mathrm{adj}}{<}10^{-3}$). $R_5$ (semantic-weighted) pushes pragmatic quality even higher ($\Delta{=}+0.038$, $p_{\mathrm{adj}}{<}10^{-3}$) but paradoxically \emph{reduces} semantic quality ($\Delta{=}-0.019$, $p_{\mathrm{adj}}{=}0.017$), the very dimension it was designed to optimize. $R_3$ (syntax-weighted) produces no significant syntactic improvement ($\Delta{=}+0.001$, n.s.) while marginally improving pragmatic quality.

Figure~\ref{fig:rq2_weights} (bottom row) shows that Qwen's weighted configurations produce compressed syntactic distributions near 0.81--0.82 while pragmatic distributions shift upward and tighten. This reveals a recurring pattern: for Qwen, \emph{all weighted rewards converge toward higher pragmatic quality at the expense of syntactic quality, regardless of which dimension is nominally emphasized}. The direction of the trade-off is consistent even when the weights are designed to pull in the opposite direction, suggesting that the optimization landscape has a natural gradient toward simplification that dominates the intended weighting signal.

\paragraph{Cross-model comparison.}
The contrast between Llama's complete weighting failure and Qwen's partial steering suggests that the effectiveness of reward weighting is architecture-dependent. Llama3.1, as the smaller model (8B), may occupy a more constrained policy space where the SFT initialization places it near a single mode; weighted rewards push the policy away without providing a path to a better one. Qwen2.5 (14B) appears to maintain a richer landscape with multiple accessible configurations, allowing weighted rewards to steer, albeit not always in the intended direction.

\begin{figure*}[t]
    \centering
    \includegraphics[width=\textwidth]{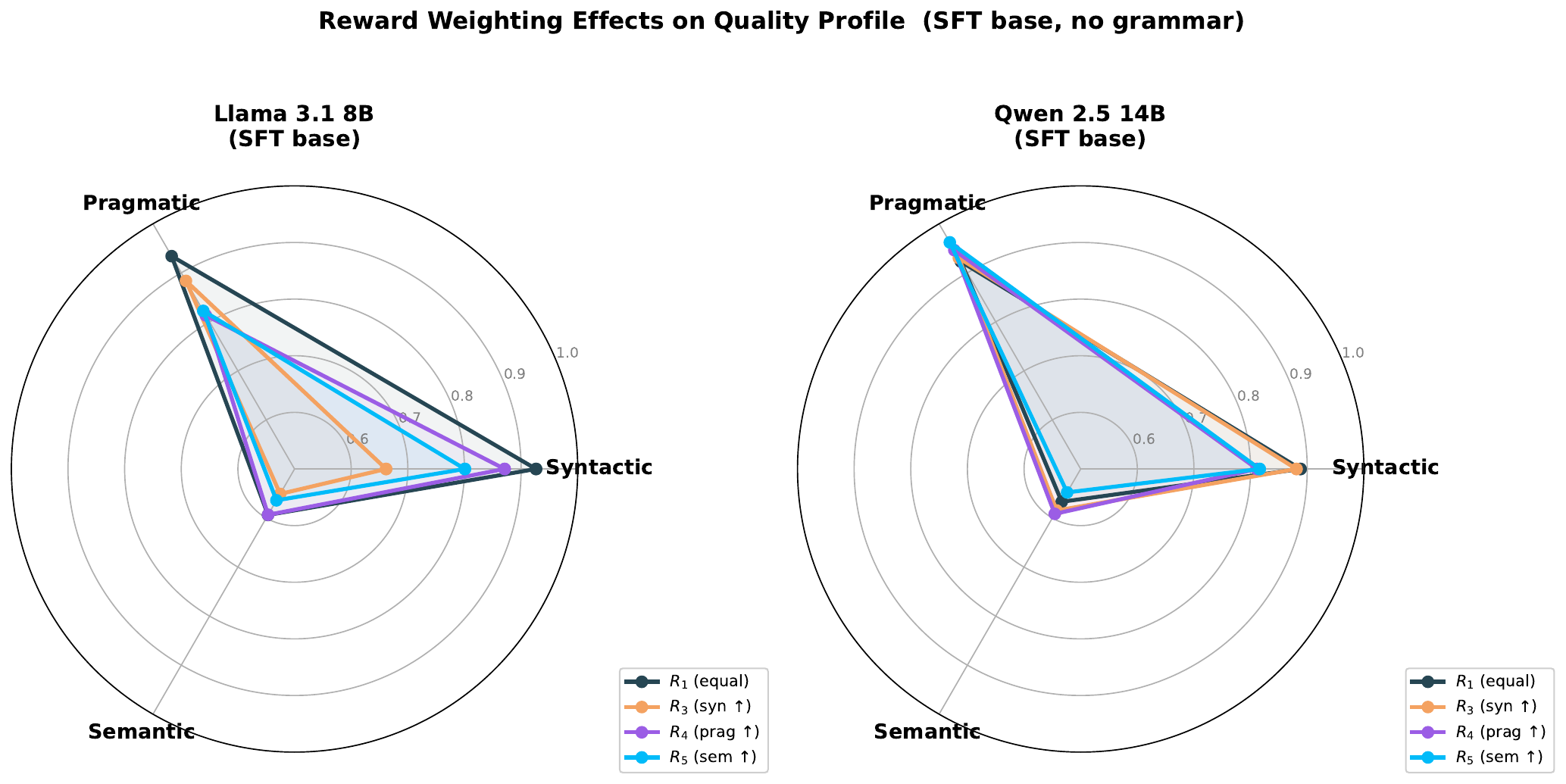}
    \caption{Radar charts comparing quality profiles of reward weighting schemes (SFT base, no grammar). For Llama (left), $R_1$ (equal) produces the largest balanced triangle. For Qwen (right), weighted variants extend the pragmatic axis while compressing the syntactic axis, regardless of which dimension receives the highest weight.}
    \label{fig:radar}
\end{figure*}

Figure~\ref{fig:radar} summarizes the quality profiles compactly. For Llama,~$R_1$ produces the largest balanced triangle; all weighted variants produce smaller, distorted shapes. For Qwen, the triangles overlap more, but the deformation pattern (pragmatic axis extending while syntactic axis contracts) is consistent across weighted configurations.

\subsubsection{Effect of the Invalidity Penalty}
\label{sec:rq2_penalty}

The comparison between~$R_1$ ($p{=}{-1}$) and~$R_2$ ($p{=}0$) isolates the effect of the negative penalty for invalid outputs, with all other parameters held constant. 
%The results reveal a striking divergence between the two model families.

\paragraph{Llama 3.1 8B.}
The penalty has no statistically significant effect on any quality dimension: $\Delta_{\mathrm{syn}}{=}+0.011$ ($p_{\mathrm{adj}}{=}0.35$), $\Delta_{\mathrm{pra}}{=}+0.002$ ($p_{\mathrm{adj}}{=}1.00$), $\Delta_{\mathrm{sem}}{=}-0.001$ ($p_{\mathrm{adj}}{=}1.00$). The distributions under~$R_1$ and~$R_2$ are nearly superimposed. This result is itself informative: it indicates that Llama's SFT-initialized policy already occupies a region of the output space where invalid outputs are rare, rendering the penalty signal redundant. The model has already ``solved'' the validity problem through SFT, and the penalty adds no further optimization pressure.

\paragraph{Qwen 2.5 14B.}
For Qwen, the penalty effect is large, highly significant, and reveals a qualitatively different optimization behavior. Removing the penalty reshapes the optimization trajectory across all dimensions: syntactic quality drops from 0.888 to 0.750 ($p_{\mathrm{adj}}{<}10^{-4}$), pragmatic quality rises from 0.923 to 0.964 ($p_{\mathrm{adj}}{<}10^{-4}$), and semantic quality decreases from 0.567 to 0.547 ($p_{\mathrm{adj}}{=}0.003$).

\begin{figure*}[t]
    \centering
    \includegraphics[width=\textwidth]{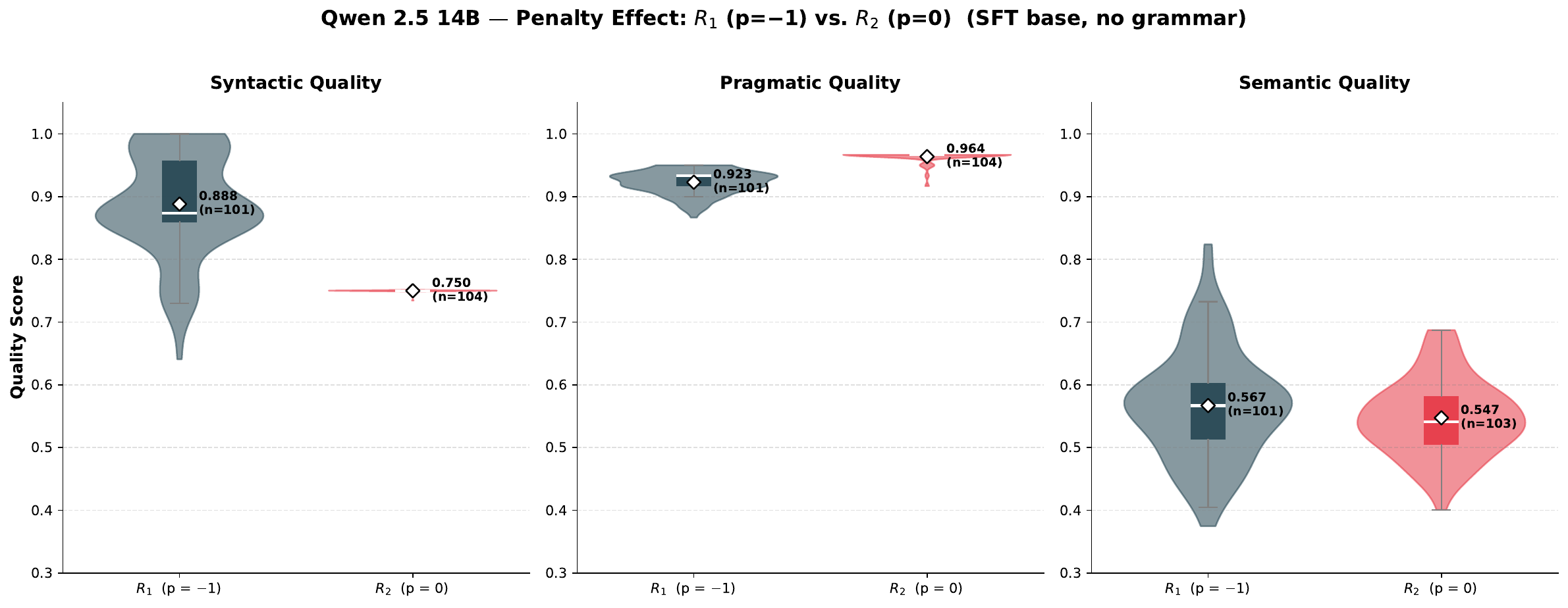}
    \caption{RQ2: Penalty effect on Qwen~2.5 14B quality distributions ($R_1$: $p{=}{-1}$ vs.\ $R_2$: $p{=}0$, SFT base, no grammar). Without the penalty ($R_2$, red), the syntactic distribution collapses to a near-uniform band at~0.750 (std${=}$0.002), while the pragmatic distribution forms a razor-thin spike at~0.964. The penalty ($R_1$, dark) preserves distributional diversity across all dimensions. For Llama (not shown), the two conditions are statistically indistinguishable.}
    \label{fig:penalty}
\end{figure*}

Figure~\ref{fig:penalty} reveals that under~$R_2$, the syntactic quality distribution collapses into an extremely narrow band centered at 0.750 with a standard deviation of only~0.002; the model generates essentially identical syntactic structures for every input. Meanwhile, the pragmatic distribution becomes a razor-thin spike at 0.964. This near-uniform output behavior stands in sharp contrast to~$R_1$, which maintains diverse distributions across all dimensions. The model under~$R_2$ has converged to what can be characterized as a single structural template: a simple, compact process model that satisfies approximately 12 of 16 syntactic rules and achieves near-perfect pragmatic scores through its minimality, applied uniformly regardless of the complexity of the input process description.

This template convergence does not occur with the penalty ($R_1$). The negative penalty for invalid outputs evidently plays a role beyond simply discouraging invalidity. It appears to act as an implicit regularizer that maintains structural diversity in the model's output. Different process descriptions require different structural solutions. A policy that must maintain validity across this diversity cannot afford to converge to a single template. The penalty thus preserves distributional variety as a side effect of enforcing broad validity.

The model-dependent nature of this finding suggests that the penalty's effectiveness depends on the base model's prior disposition toward producing valid outputs. When validity is already reliable (Llama~SFT), the penalty is inert. When the model's validity behavior is more variable (Qwen), the penalty becomes a critical design choice that determines whether the optimization produces a diverse, high-quality policy or a collapsed, template-based one.

% ============================================================
\subsection{RQ3: How Does Base Model Initialization Affect RL Optimization?}
\label{sec:rq3}

To assess the importance of SFT initialization for RL training, we compare models trained with GSPO from an SFT-initialized base against models trained directly from the BPM-unadapted instruction-tuned base, both using~$R_1$ and no grammar enforcement. Statistical comparisons use paired permutation tests with Bonferroni correction ($m{=}3$).

\subsubsection{Llama3.1 8B}

For Llama, SFT initialization produces large, significant advantages across all dimensions: syntactic quality ($\Delta{=}+0.262$, $p_{\mathrm{adj}}{<}10^{-4}$), pragmatic quality ($\Delta{=}+0.049$, $p_{\mathrm{adj}}{<}10^{-4}$), and semantic quality ($\Delta{=}+0.036$, $p_{\mathrm{adj}}{<}10^{-4}$). The syntactic quality gap of~0.262 is comparable in magnitude to the~$R_3$ weighting collapse ($\Delta{=}0.264$ in Section~\ref{sec:rq2_weights}), making these two effects (initialization failure and weighting collapse) the largest observed in this study, both exceeding 0.25 points. That the two largest effects both involve a form of policy collapse (convergence to a narrow, low-quality output mode) suggests a common underlying mechanism that is explored further in Section~\ref{sec:discussion}.

\begin{figure*}[t]
    \centering
    \includegraphics[width=\textwidth]{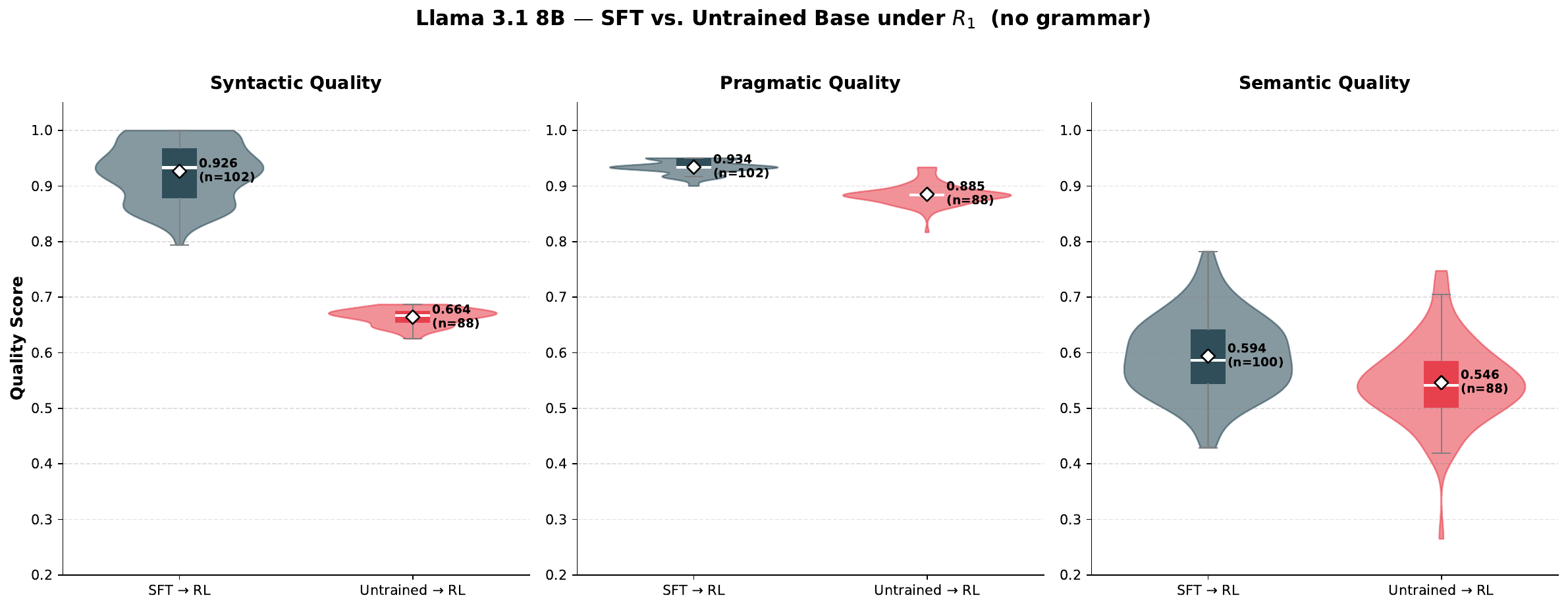}
    \caption{RQ3: Effect of base model initialization on RL outcomes under~$R_1$ for Llama~3.1 8B (no grammar). The SFT-initialized base (dark) produces high syntactic quality (0.926), while the BPM-unadapted base (red), i.e., the original instruction-tuned model without BPM-specific SFT, collapses to~0.664, a gap of 0.262, one of the two largest effects in this study. Pragmatic quality shows a smaller but significant gap. Semantic quality benefits from SFT but the difference is moderate. For Qwen (not shown), SFT provides no significant advantage on syntactic or pragmatic quality, and the BPM-unadapted base achieves significantly higher semantic quality (0.601 vs.\ 0.567).}
    \label{fig:rq3}
\end{figure*}

Figure~\ref{fig:rq3} provides essential distributional context. The BPM-unadapted-base violin for syntactic quality forms a tight cluster at 0.664, well below even the worst SFT-based RL configurations. This indicates that the instruction-tuned Llama model, without BPM-specific SFT, converges to a low-quality equilibrium under GSPO: the base model's output distribution is sufficiently far from the target format that the sampled candidates are uniformly poor, and the group-relative ranking within each generation group provides only noise rather than informative gradients for improvement. The SFT stage is essential to bring the policy within a region of the output space where quality-relevant distinctions between candidates become detectable.

The pragmatic quality gap is smaller but still highly significant ($\Delta{=}+0.049$). Interestingly, the BPM-unadapted base achieves 0.885 pragmatic quality, already a reasonable level, while falling drastically short on syntax. This asymmetry confirms that pragmatic quality (which rewards simplicity) is achievable even without format-specific training, whereas syntactic quality (which requires rule-specific structural correctness) depends critically on initialization.

\subsubsection{Qwen2.5 14B}

The pattern for Qwen reverses on the semantic dimension. SFT initialization provides no significant advantage for syntactic quality ($\Delta{=}+0.017$, $p_{\mathrm{adj}}{=}0.16$) or pragmatic quality ($\Delta{=}+0.003$, $p_{\mathrm{adj}}{=}0.71$), but the BPM-unadapted base \emph{significantly outperforms} the SFT base on semantic quality ($\Delta{=}-0.032$, $p_{\mathrm{adj}}{<}10^{-4}$).

This reversal has a clear implication: for Qwen, SFT initialization is associated with lower semantic quality after RL. The Qwen architecture appears to handle the target output format sufficiently well from its instruction-tuned state, likely due to its larger parameter count and broader training exposure to structured and JSON-formatted content, so SFT initialization provides no structural advantage. Meanwhile, the SFT stage may anchor the policy toward the specific output patterns present in the BPM training data, potentially constraining the model from exploring alternative process model structures during RL that would better capture the semantic content of novel process descriptions. The BPM-unadapted base, unconstrained by SFT-imposed patterns, retains greater flexibility to produce semantically diverse outputs.

This finding carries a direct practical implication: the standard two-stage pipeline (SFT~$\rightarrow$~RL) should not be assumed as universally beneficial. For models whose instruction-tuned capabilities already align with the target task's structural requirements, skipping BPM-specific SFT may yield better semantic outcomes under RL, while also saving the computational cost of the SFT stage. A preliminary zero-shot or direct-RL pilot evaluation may help assess whether BPM-specific SFT is necessary before committing to the full training pipeline.

\subsubsection{Summary}

The necessity and effect of SFT initialization is strongly architecture-dependent. For Llama, SFT is indispensable: without it, RL cannot produce syntactically competent outputs. For Qwen, SFT is largely redundant for structural quality and is associated with lower semantic quality after RL. This divergence is consistent with the broader pattern observed throughout the experiments: the same design choice (reward weighting, penalty regime, initialization strategy) can produce qualitatively different outcomes depending on the model architecture, underscoring the importance of empirical evaluation over universal design assumptions.

\section{Discussion}
\label{sec:discussion}

%This section interprets the experimental findings, connects them to broader principles in reinforcement learning for language models, compares with related work, and addresses limitations. We organize the discussion around four thematic subsections followed by limitations and future work.

%--------------------------------------------------------------
\subsection{Optimization Dynamics and Reward Design}
\label{sec:disc_optim}

The most robust finding across all experiments is that the three quality dimensions respond differently to RL optimization: pragmatic quality improves readily and consistently ($\Delta > 0.11$ for both models), syntactic quality improves significantly but model-dependently (strongly for Llama, modestly for Qwen), and, under the equal-weight SFT-initialized condition used for RQ1, semantic quality remains largely unchanged, showing a small significant improvement for Llama but no significant effect for Qwen. This ordering reflects what we term the \emph{optimization accessibility} of each dimension, the ease with which a policy can improve its score through behavioral change. Pragmatic quality is the most accessible because it rewards simpler, more compact outputs, a direction always available in the policy space without requiring domain-specific knowledge. Syntactic quality requires learning specific BPMN structural rules, a more targeted adjustment. Semantic quality depends on correspondence with an external reference that is available only through the post-generation reward signal, not during generation itself; this makes the optimization signal more indirect and less readily exploitable than syntactic or pragmatic feedback. Importantly, the equal-weight reward achieves structural and comprehensibility improvements without sacrificing semantic fidelity, an outcome that is not guaranteed, as the aggressive penalty-free configuration ($R_2$) demonstrates. Practitioners should therefore expect large pragmatic gains, moderate syntactic gains (particularly for weaker baselines), and minimal semantic effects from the approach presented here, provided that the reward function maintains balanced dimensional pressure.

This accessibility gradient also explains the most counter-intuitive result in the study: the failure of weighted rewards. $R_3$ (syntax-weighted, $3{:}1{:}1$) produces the \emph{lowest} syntactic quality for Llama (0.662), and~$R_5$ (semantic-weighted) reduces semantic quality for both models. We attribute this to two complementary mechanisms. First, the $3{:}1{:}1$ weighting reduces the multi-dimensional reward to near-single-objective optimization, making GSPO's group-relative advantage computation dependent on variation along a single axis. When candidates show limited variation on the dominant dimension, as is likely for an SFT-initialized policy, the advantage signal becomes noisy and the policy drifts erratically rather than improving. 
Second, the three quality dimensions share structural underpinnings in BPMN generation: syntactic correctness, pragmatic complexity, and semantic fidelity are all properties of the same process graph. Focusing on optimizing one dimension disrupts the structural coherence that all three depend on. Equal weighting avoids this behavior by penalizing changes that improve one dimension at the expense of others.

This pattern likely reflects underlying differences in the geometry of each model's policy space. For Llama, the landscape appears relatively narrow, where SFT pushes the model toward a single dominant mode, and optimization pressure away from that region finds no stable alternatives nearby, resulting in full policy collapse.
Qwen seems to operate in a more flexible landscape, with multiple accessible local optima that allow weighted rewards to steer behavior without destabilizing the policy entirely. Even so, optimization does not primarily improve along the intended behavioral dimension — gains instead appear to come from easier or less constrained directions in the policy space. 
These findings echo the multi-objective optimization literature where naive scalarization produces Pareto-dominated solutions when objectives are correlated~\cite{deb2002fast}, and provide empirical evidence that this concern manifests concretely in RL for structured generation. Unless strong evidence supports a specific weighting scheme, equal weighting should be the default.

%--------------------------------------------------------------
\subsection{Template Collapse and the Invalidity Penalty}
\label{sec:disc_collapse}

Qwen's behavior under~$R_2$ (no penalty), converging to a near-uniform output template with syntactic standard deviation of 0.002, illustrates how reward optimization can produce degenerate solutions that achieve high objective values while violating the intent. The model discovers a single structure achieving 0.750 syntactic quality and 0.964 pragmatic quality through minimality, then applies it identically to all inputs. This resembles reward hacking in the RL alignment literature~\cite{amodei2016concrete}, but with an important distinction: the template strategy achieves high reward under the defined objective. The problem lies in the reward \emph{design}, specifically the absence of a diversity-preserving mechanism.

The negative penalty ($p{=}{-1}$) mitigates this collapse through an indirect mechanism. By strongly penalizing invalid outputs, it forces the policy to maintain validity across diverse inputs. Since different process descriptions require different structural solutions for valid modeling, maintaining broad validity implicitly requires maintaining a diverse structural repertoire. The penalty thus does not explicitly reward variation, but its interaction with the diverse evaluation set reduces the risk of convergence to a single template.

The model-dependent nature of this effect is revealing. Llama's SFT-initialized policy operates deep within the high-validity region, where the penalty has little measurable effect. Qwen's policy explores closer to the validity boundary, making the penalty an active constraint. This means that if a model achieves near-perfect validity under SFT alone, the penalty may be less consequential; if validity is imperfect, the penalty appears important for reducing the risk of degenerate convergence.

%--------------------------------------------------------------
\subsection{Initialization and the Exploration--Exploitation Trade-Off}
\label{sec:disc_init}

The RQ3 results (SFT indispensable for Llama under our training configuration, largely redundant and associated with lower semantic quality for Qwen) reflect different positions on the exploration--exploitation spectrum. SFT initialization is an exploitation-heavy strategy that places the policy near a known good region of the output space but constrains subsequent exploration. The BPM-unadapted base provides broader exploration but risks starting too far from productive regions of the policy space.

For Llama, the BPM-unadapted starting point is too distant: the instruction-tuned base model's output distribution lacks the structural regularity for process model generation, candidates are uniformly poor, and GSPO's relative ranking provides only noise. In our experiments, SFT is essential to bring the policy within a region where quality-relevant distinctions become detectable. For Qwen, the instruction-tuned distribution already approximates the target format, likely due to larger scale and broader training data, so SFT provides marginal structural benefit while constraining semantic exploration. The BPM-unadapted base retains flexibility to generate structurally novel outputs, explaining the semantic quality reversal (0.601 vs.\ 0.567).

The parallel between this initialization-driven policy collapse (Llama BPM-unadapted at 0.664 syntactic) and the weighting-driven policy collapse from Section~\ref{sec:disc_optim} (Llama~$R_3$ at 0.662 syntactic) deserves emphasis. Both involve convergence to a narrow, low-diversity output mode from which GSPO cannot escape, a fundamental limitation of group-relative optimization that emerges whenever all candidates within a group are similarly poor. This shared failure mode suggests that ensuring the policy operates within a productive region of the output space is a precondition for effective GSPO training, whether that assurance comes through SFT initialization, appropriate reward design, or both.

Two supplementary observations reinforce this interpretation. First, in an exploratory comparison, $R_{\mathrm{avg}}$ and~$R_1$, which are mathematically equivalent in scalar reward value but implemented through separate reward pipelines, produce different outcomes for Llama ($\Delta{=}+0.059$ syntactic, $p_{\mathrm{adj}} < 2 \times 10^{-5}$, favoring~$R_1$) but not Qwen. Because both configurations are equivalent in expectation, we interpret this difference cautiously: it may reflect implementation-level effects, stochastic training variability, or sensitivity of smaller models to minor reward-pipeline differences rather than a principled advantage of reward decomposition. Second, for the BPM-unadapted Qwen base, $R_5$ (semantic-weighted) achieves the highest syntactic quality in the entire study (0.928), exceeding all SFT-based configurations, a concrete example of the exploration dividend available when initialization is appropriately matched to the model's capabilities. We flag both observations as preliminary findings that warrant dedicated investigation.

%--------------------------------------------------------------
\subsection{Grammar-Constrained Decoding and Comparison with Related Work}
\label{sec:disc_related}

In supplementary analyses of the grammar condition, which we summarize here due to space constraints, grammar-constrained decoding at inference time produced no clear effect for well-optimized configurations ($R_1$ with SFT base); the learned policy already generates structurally well-formed outputs, making grammar enforcement rather redundant. For weaker configurations (weighted rewards, BPM-unadapted bases), grammar provides modest validity improvements but does not address quality deficiencies rooted in suboptimal reward design or initialization. Grammar is thus complementary for underperforming models but not a substitute for appropriate reward engineering.

The most directly comparable prior work is \citet{berti2025specializing}, who apply RL with verifiable rewards to improve LLM-based process modeling. Our approach differs in using GSPO (sequence-level group-relative ranking) rather than GRPO (token-level), in systematically investigating reward composition rather than using a single formulation, and in evaluating with the BEF4LLM framework's 38 metrics across three theoretically grounded quality dimensions~\cite{lauer2026bef4llm}. Our findings complement theirs in confirming that RL improves process model generation, but add the critical insight that \emph{how the reward is designed matters as much as whether RL is applied}. The weighting, penalty, and initialization analyses demonstrate that seemingly minor design choices produce effects as large as the RL-vs-SFT improvement itself, a level of analysis absent from prior work that underscores the need for systematic empirical evaluation rather than ad-hoc reward engineering.

%--------------------------------------------------------------
\subsection{Limitations and Future Work}
\label{sec:limitations}

Several limitations apply to this work. The evaluation uses 105 held-out samples with a single generation per sample, which provides adequate power for the large effects observed but limits sensitivity to small effects and prevents estimation of within-configuration variance. The weighted reward variants use a fixed $3{:}1{:}1$ ratio; the complete failure for Llama does not necessarily imply that gentler imbalances would also fail. The analysis covers two LLM families differing in both architecture and scale, which provides useful diversity but does not allow disentangling these factors. Semantic quality is measured against ground-truth references, potentially underestimating models that generate valid but structurally novel alternatives. Finally, single-epoch GSPO training and the use of LoRA mitigate but do not eliminate the risks of reward overfitting and catastrophic forgetting, respectively; neither was explicitly tested.

Several directions emerge from these limitations. \emph{Adaptive reward weighting}, dynamically adjusting dimensional weights during training based on the current quality profile, could address the weighting failure by preventing any single dimension from dominating the gradient while still enabling targeted improvement. \emph{Scaling and generalization} to larger models and other structured generation domains (UML diagrams, database schemas, workflow specifications) would test whether the findings are specific to the BPMN setting or reflect broader properties of multi-dimensional RL for structured outputs. \emph{Semantic reward enrichment} through embedding-based similarity measures or language-model-based semantic judgments could provide the targeted gradient information that metric-based semantic rewards currently lack. Finally, \emph{multi-epoch RL with regularization}, combining extended GSPO training with KL-divergence constraints or validation-based early stopping, could unlock further quality gains while maintaining output diversity.
\section{Conclusion}
\label{sec:conclusion}

This paper presented a systematic investigation of RL with domain-grounded reward signals for improving LLM-based business process model generation. Using GSPO with reward functions derived from the BEF4LLM quality framework, we trained two open-source LLM families (Llama~3.1 8B and Qwen~2.5 14B) under 48 experimental configurations that vary reward weighting, invalidity penalty, base model initialization, and grammar-constrained decoding. The configurations were evaluated on 105 held-out process descriptions using paired permutation tests with Bonferroni correction across three quality dimensions: syntactic, pragmatic, and semantic quality.

The findings address three research questions and yield both practical recommendations and broader insights for reward function design in structured generation.
Regarding RQ1, GSPO with equal-weight rewards significantly improves pragmatic quality for both model families (by more than 0.11 points) and syntactic quality (strongly for Llama, modestly for Qwen), while preserving semantic fidelity. Beyond the mean improvement, GSPO reduces output variability by more than a factor of six for pragmatic quality among valid generated outputs, producing substantially more consistent and deployment-ready process models compared to SFT-only baselines.

Regarding RQ2, equal reward weighting is the most robust default in our experiments for multi-dimensional quality optimization. Targeted weighting (3:1:1 emphasis on a single dimension) consistently fails to improve the intended dimension and, in the most extreme case, causes the model to collapse into a narrow low-quality mode with a syntactic quality drop of 0.264 points. The invalidity penalty plays a model-dependent role: it has little measurable effect when the SFT-initialized policy already produces valid outputs reliably (Llama), but is crucial for preventing template collapse when the model's validity behavior is more variable (Qwen). These results demonstrate that the composition of the reward function is not a secondary design choice but a primary determinant of optimization outcomes whose effects can be as large as the decision to apply RL in the first place.

Regarding RQ3, the necessity of SFT initialization is architecture-dependent. For Llama, SFT is indispensable under our training configuration, providing the structural foundation without which RL converges to a low-quality equilibrium. For Qwen, SFT is largely redundant for syntactic and pragmatic quality and is associated with lower semantic quality after RL, suggesting that it may constrain the policy space and limit semantically richer alternatives. This finding challenges the assumption that the standard SFT-then-RL pipeline is universally optimal and suggests that initialization strategy should be co-designed with the base model's task-relevant prior capabilities.

Three implications extend beyond the BPMN domain. First, for any structured generation task where quality is assessed along multiple automated dimensions, our results caution against naive reward weighting and recommend equal weighting as the starting point. Second, the observation that a penalty for invalid outputs can function as an implicit diversity regularizer suggests that reward components designed for one purpose (validity enforcement) may have important secondary effects (output diversity) that should be anticipated in reward engineering. Third, the strong interaction between reward design, model architecture, and initialization strategy implies that transfer of reward configurations across model families is unreliable, and practitioners should treat these as jointly tunable rather than independently settable parameters.

Future work should explore adaptive reward weighting that adjusts dimensional emphasis during training based on the current quality profile, semantic reward enrichment through embedding-based or LLM-based evaluation to address the optimization inaccessibility of semantic quality, and multi-epoch RL training with regularization mechanisms to prevent reward overfitting while unlocking further quality gains. Extending the analysis to larger models, additional LLM architectures, and other structured generation domains such as UML diagrams, database schemas, and workflow specifications would test the generality of the findings reported here.

%\appendix

\paragraph*{\textbf{Acknowledgment}}
This work work was partly conducted within the project KICoPro (Grant 01IS24053C), funded by the Federal Ministry of Research, Technology and Space (BMFTR).

\paragraph*{\textbf{Declaration of generative A}}

During the preparation of this work the author(s) used Claude Opus 4.6 in order to to enhance language clarity, coherence, and rephrasing. After using this tool/service, the author(s) reviewed and edited the content as needed and take(s) full responsibility for the content of the published article.

\printcredits

\bibliographystyle{cas-model2-names}

\bibliography{bib}



\end{document}